\documentclass[journal]{IEEEtran}
\usepackage{graphicx}

\usepackage{cite}
\usepackage{picinpar}
\usepackage{amsmath}
\usepackage{url}
\usepackage{flushend}
\usepackage{colortbl}
\usepackage{soul}
\usepackage{multirow}
\usepackage{pifont}
\usepackage{color}
\usepackage{alltt}
\usepackage[hidelinks]{hyperref}
\usepackage{enumerate}
\usepackage[mode=text]{siunitx}
\usepackage{epstopdf}
\usepackage{pbox}
\usepackage{hhline}
\usepackage{amssymb}
\usepackage{stfloats}
\usepackage{float}

\usepackage{xcolor}
\usepackage[ruled,linesnumbered]{algorithm2e}
\usepackage{tabularx}
\usepackage{multirow}
\usepackage{booktabs}
\usepackage{enumitem}
\usepackage{epstopdf} 

\usepackage{dcolumn} 
  \newcolumntype{d}[1]{D{.}{.}{#1}}    
  \newcommand{\tabincell}[2]{\begin{tabular}{@{}#1@{}}#2\end{tabular}}

\DeclareSymbolFont{largesymbol}{OMX}{yhex}{m}{n}
\DeclareMathAccent{\Widehat}{\mathord}{largesymbol}{"62}

\title{Closed-Loop Magnetic Manipulation for Robotic Transesophageal Echocardiography}
\author{Keyu~Li$^{*}$,~\IEEEmembership{Graduate Student Member,~IEEE},
	    Yangxin~Xu$^{*}$,
        Ziqi~Zhao,
        Ang~Li,~\IEEEmembership{Graduate Student Member,~IEEE},
        and~Max~Q.-H.~Meng$^{\sharp}$,~\IEEEmembership{Fellow,~IEEE}
\thanks{This work was partially supported by National Key R\&D program of China with Grant No. 2019YFB1312400, Hong Kong RGC CRF grant C4063-18G, and Hong Kong RGC GRF grant \# 14211420 awarded to Max Q.-H. Meng.}
\thanks{K. Li and A. Li are with the Department of Electronic Engineering, The Chinese University of Hong Kong, Hong Kong (e-mail: kyli@link.cuhk.edu.hk; psw.liang@link.cuhk.edu.hk).}
\thanks{Y. Xu is with Yuanhua Robotics, Perception \& AI Technologies Ltd, Shenzhen, China (e-mail: yxxu@link.cuhk.edu.hk).}
\thanks{Z. Zhao is with the Department of Electronic and Electrical Engineering, the Southern University of Science and Technology, Shenzhen, China (e-mail: zhaozq2020@mail.sustech.edu.cn).}
\thanks{Max Q.-H. Meng is with the Department of Electronic and Electrical Engineering of the Southern University of Science and Technology in Shenzhen, China, on leave from the Department of Electronic Engineering, the Chinese University of Hong Kong, Hong Kong SAR, China, and also with the Shenzhen Research Institute of the Chinese University of Hong Kong, Shenzhen, China (e-mail: max.meng@ieee.org).}
\thanks{$^{*}$ indicates equal contribution.}
\thanks{$^{\sharp}$ Corresponding author.}}

\begin{document}

\maketitle

\begin{abstract}
This paper presents a closed-loop magnetic manipulation framework for robotic transesophageal echocardiography (TEE) acquisitions. 
Different from previous work on intracorporeal robotic ultrasound acquisitions that focus on continuum robot control, we first investigate the use of magnetic control methods for more direct, intuitive, and accurate manipulation of the distal tip of the probe. 
We modify a standard TEE probe by attaching a permanent magnet and an inertial measurement unit sensor to the probe tip and replacing the flexible gastroscope with a soft tether containing only wires for transmitting ultrasound and IMU data, and show that 6-DOF localization and 5-DOF closed-loop control of the probe can be achieved with an external permanent magnet based on the fusion of internal inertial measurement and external magnetic field sensing data. 
The proposed method does not require complex structures or motions of the actuator and the probe compared with existing magnetic manipulation methods. 
We have conducted extensive experiments to validate the effectiveness of the framework in terms of localization accuracy, update rate, workspace size, and tracking accuracy. In addition, our results obtained on a realistic cardiac tissue-mimicking phantom show that the proposed framework is applicable in real conditions and can generally meet the requirements for tele-operated TEE acquisitions. 
\end{abstract}


\section{Introduction}

\begin{figure}[t]
\centering
\includegraphics[scale=1.0,angle=0,width=0.45\textwidth]{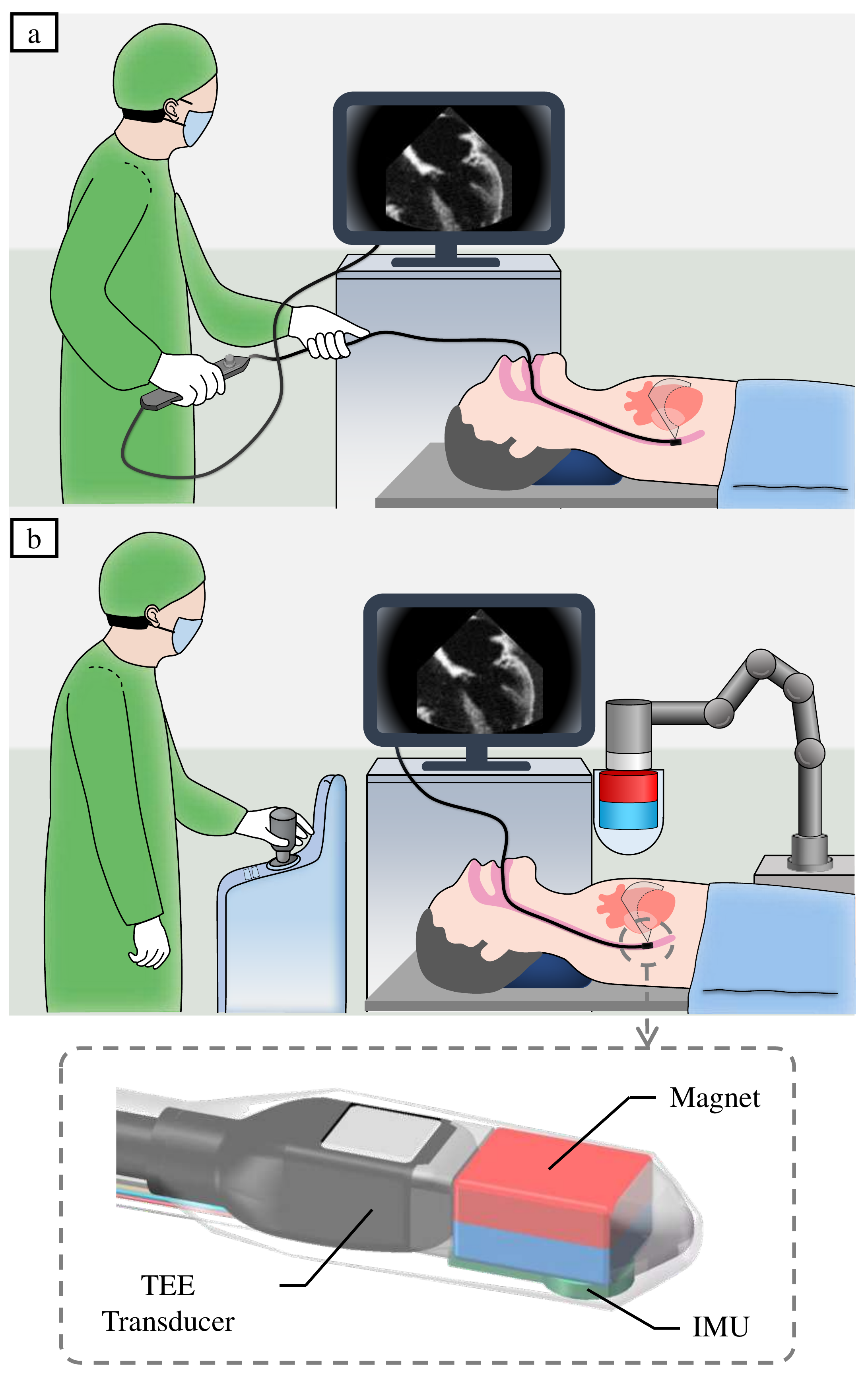}
\caption{(a) illustrates the procedure of conventional TEE acquisitions, where the echocardiographer manually grips the flexible shaft and adjusts the knobs on the probe handle to advance and steer the probe in the esophagus. (b) shows the proposed application scenario of the MA-TEE system, where the probe tip attached to a permanent magnet and an IMU sensor is manipulated by an external permanent magnet held by a robotic manipulator for tele-operated TEE acquisitions.}
\label{Fig_app_system}
\end{figure}

\IEEEPARstart{U}{ltrasound} imaging is widely used for diagnosis and treatment in many medical disciplines, providing major advantages of safety, portability, low cost, and real-time capabilities compared with other medical imaging modalities such as CT and MRI \cite{chan2011basics}. 
Transesophageal echocardiography (TEE) is a test that uses a specialized ultrasound probe placed inside the esophagus to assess heart structures \cite{khandheria1994transesophageal}. As shown in Fig. \ref{Fig_app_system}(a), in conventional TEE examinations, an experienced echocardiographer manually grips the flexible shaft and adjusts the knobs on the probe handle to advance and steer the distal tip of the probe in the esophagus \cite{hahn2013guidelines}. Since the operator can only manipulate the proximal end of the semi-rigid tube, TEE operation usually requires a long learning curve and extensive training, and the imaging quality is highly dependent on the operator \cite{bloch2017impact}. Moreover, poor image quality and repeated insertion are associated with an increased risk of complications \cite{freitas2020safety}. In addition, the close contact between the echocardiographer and the patient would increase the risk of infection under epidemic situations \cite{yang2020combating}.

Recently, a large number of robotic systems have been developed for tele-operated or autonomous ultrasound acquisitions \cite{li2021overview}. However, existing studies have mainly focused on extracorporeal applications \cite{mustafa2013robio,chatelain2017confidence,vesnet, li2021autonomous,
ning2021autonomic, li2021image}, which typically use a robotic manipulator to directly maneuver the probe on the skin surface to scan the region of interest. Only a few studies have investigated the robotic ultrasound acquisitions in intracorporeal applications \cite{norton2019intelligent, loschak2016algorithms,wang2016robotic, wang2016design}. 
Wang \textit{et al.} \cite{wang2016robotic, wang2016design} developed the first robotic systems to automatically hold the TEE probe handle to manipulate a standard TEE probe with 4 DOFs. The probe tip was tracked based on the registration between intraoperative ultrasound images and preoperative MRI data of the patient \cite{wang2016robotic}.  
However, since the system still used a standard TEE probe with a flexible gastroscope, it cannot overcome the inherent limitations such as the lack of intuitiveness, indirect and complex control of the probe tip, and the risk of patient injury due to tissue stretching as the semi-rigid endoscope is pushed through the esophagus. Also, the image registration-based method for probe localization was reported to have a limited computational speed and reliability \cite{wang2016robotic}. 
In addition, there is no mechanism to adjust the contact force between the ultrasound transducer and tissue due to the use of a standard TEE probe. 

Magnetic locomotion has become a promising technology to wirelessly manipulate a medical device in the human body without requiring mechanical linkage between the actuator and the device, which has shown great promise to overcome the limitations of standard endoscopic devices \cite{abbott2020magnetic, sliker2014flexible, xu2022autonomous, xu2022adaptive}.
In \cite{norton2019intelligent}, Norton \textit{et al.} presented the first demonstration of US acquisitions in the bowel using a magnetically manipulated capsule, in which a magnet, the miniature US transducers, and several magnetic field sensors were embedded.

In this work, we are committed to presenting a more straightforward approach to robotic TEE manipulation based on magnetic methods to allow for more direct, intuitive and accurate control of the TEE probe tip, thereby improving ease of use and patient safety for the targeted application. 
The proposed application scenario of the magnetically actuated TEE (MA-TEE) system is shown in Fig. \ref{Fig_app_system}(c), where the internal probe is manipulated by an external permanent magnet held by a robotic manipulator. 
By attaching a single permanent magnet and an inertial measurement unit (IMU) sensor to the probe tip and replacing the flexible gastroscope with a soft tether that contains only wires for transmitting ultrasound and IMU data, the modified TEE probe can be viewed as a tethered capsule, and we show that 6-DOF localization and 5-DOF control of the probe can be achieved using an external and internal sensor fusion based approach. 
The proposed system can allow the user to focus on the ultrasound images on the screen and directly instruct the desired movement of the probe tip, and the actuator magnet will be automatically adjusted to realize the desired motion of the probe. 

To our knowledge, this is the first robotic TEE system based on magnetic control methods. Compared with previous studies on intracorporeal robotic ultrasound acquisitions \cite{loschak2016algorithms, wang2016robotic, wang2016design}, the method proposed in this paper does not require complex modelling of the continuum robot kinematics and can directly control the distal end of the probe, which provides a simpler and more intuitive method for robotic TEE manipulation. Moreover, instead of tracking the probe based on image registration, our method can estimate the 6-DOF pose of the probe tip in real time based on the fusion of internal inertial sensing data and external magnetic field sensing data, which can realize more accurate and efficient localization of the probe tip and provide additional diagnostic information for the echocardiographer. In addition, the modified TEE probe with a string-like soft tether instead of a standard flexible endoscope may help improve patient safety during the manipulation \cite{sliker2014flexible}.
Compared with existing magnetic control methods (e.g. \cite{norton2019intelligent, popek2017first, xu2020novelsystem}), our method can greatly simplify the structures of the actuator and the probe, and can achieve 6-DOF localization and 5-DOF control without requiring the magnets to perform specific motions.

This paper is organized as follows. We first introduce the related work in Section~II. Nomenclature in this paper is presented in Section~III. The details of the proposed closed-loop manipulation framework are introduced in Section~IV. Then, our experimental results are presented in Section~V, before conclusions are drawn in Section~VI.

\section{Related Work}
In recent years, various magnetic actuation systems have been proposed for remote control and steering of catheters and guidewires for vascular interventions \cite{hwang2020review, ali2016steerable, heunis2021design, kim2019ferromagnetic}, which have demonstrated potential benefits of reduced procedural time and improved access to hard-to-reach regions. While our work shares some similarities with these systems, the major difference is that the localization of the magnetic catheters and guidewires is usually performed by observation using an external 3D imaging system (e.g., fluoroscopy), which is not available in intracorporeal ultrasound imaging applications.

Several groups have explored the simultaneous magnetic actuation and localization for magnetic capsule endoscopy using electromagnet coils \cite{son2017magnetically, guitron2017autonomous, 9730063} and permanent magnets \cite{popek2017first, xu2020novelsystem, xu2022autonomous, xu2022adaptive} as the magnetic field sources.
Compared with the systems that use electromagnetic coils to actuate the capsule, the permanent-magnet-based systems are generally more compact, cost-effective and energy-efficient, and often provide a larger workspace. Therefore, in this work, we use a single permanent magnet as the external actuator, which is maneuvered by a robotic manipulator, similar to \cite{xu2020novelsystem, xu2022adaptive}.

Localization is essential to achieve closed-loop magnetic control of the internal capsule robot. Some researchers proposed to use magnetic field sensors placed inside the capsule to measure the magnetic field of the external magnet for capsule localization \cite{popek2013localization, popek2016six, taddese2018enhanced}. However, these methods require up to six sensors to be embedded in the capsule with a specific structure to prevent saturation of the sensors, which is challenging in practice and would increase the size of the internal device. Moreover, to address the localization singularity of the internal-sensor-based methods, these systems either use an additional coil in the actuator \cite{taddese2018enhanced} or require the magnets to make specific motions during the capsule localization \cite{popek2016six}, which further increase the complexity of the overall approach. 
As an alternative, some researchers proposed to locate the capsule using a magnetic field sensor array placed external to the capsule, so that the capsule can be simplified to contain only a permanent magnet \cite{hu2005efficient, xu2018free, xu2020improved, xu2020novelsystem}. However, these methods can only solve 5-DOF pose of a static capsule and have a limited workspace size.
Recently, we presented a localization method that combines the use of an external magnetic field sensor array and an internal IMU sensor inside the capsule to estimate the full 6-DOF pose of a magnetic capsule robot \cite{li2022external}. The hybrid approach does not require complex structures of the actuator and the capsule, and can avoid the need for the magnets to make specific motions.

In this paper, we move one step further to develop a closed-loop magnetic manipulation framework that can allow 5-DOF control of a capsule robot (TEE probe tip) based on external and internal sensor fusion. Compared with the aforementioned methods, our method only use a single permanent magnet in the actuator, and the internal control only include a magnet and an IMU, which can greatly simplify the structures of the actuator and the capsule for easier application in clinical use. Moreover, the proposed method can achieve full 6-DOF localization and 5-DOF control of the internal robot without requiring the magnets to perform specific motions during the localization. Finally, we present the first demonstration of magnetically actuated robotic TEE acquisitions in a tissue-mimicking phantom to validate the feasibility of the framework.

\section{Nomenclature}

Throughout this paper, scalars are denoted by normal fonts (e.g., $\mu_0$), vectors are denoted by bold lowercase fonts (e.g., $\textbf{p}_c$), and matrices are denoted by bold uppercase fonts (e.g. $\mathbf{K}$). $\mathbf{R}_x (\alpha)$ denotes the rotation matrix that rotates a vector by an angle $\alpha$ about the $x$-axis in three dimensions. A hat over a symbol (e.g., $\widehat{\mathbf{m}}_c$) indicates the unit vector of the original vector (e.g., $\mathbf{m}_c$).  A tilde over a symbol (e.g., $\widetilde{\mathbf{b}}_{i}$) represents the measurement of that variable. A bar over a symbol (e.g., $\bar{\gamma}$) denotes the \textit{a priori} prediction. 

We follow the convention of previous works to use ``capsule" to refer to the magnet inside the body to be controlled, and ``actuator" to refer to the magnet outside the body to provide the actuating magnetic fields. The subscripts $c$ and $a$ are used to represent the ``capsule" and the ``actuator", respectively, and the subscript $d$ means``desired". $\mathbb{R}$ denotes the real number set. $\mathbf{x} \sim \mathcal{N}(\boldsymbol{\mu}, \mathbf{\Sigma})$ means a random vector $\mathbf{x}$ follows a normal distribution with mean vector $\boldsymbol{\mu}$ and covariance matrix $\mathbf{\Sigma}$.
 $\mathbf{I}_{k}$ indicates a $k \times k$ identity matrix.
\section{Methodology}

\subsection{Problem Formulation}

\begin{figure}[t]
\centering
\includegraphics[scale=1.0,angle=0,width=0.48\textwidth]{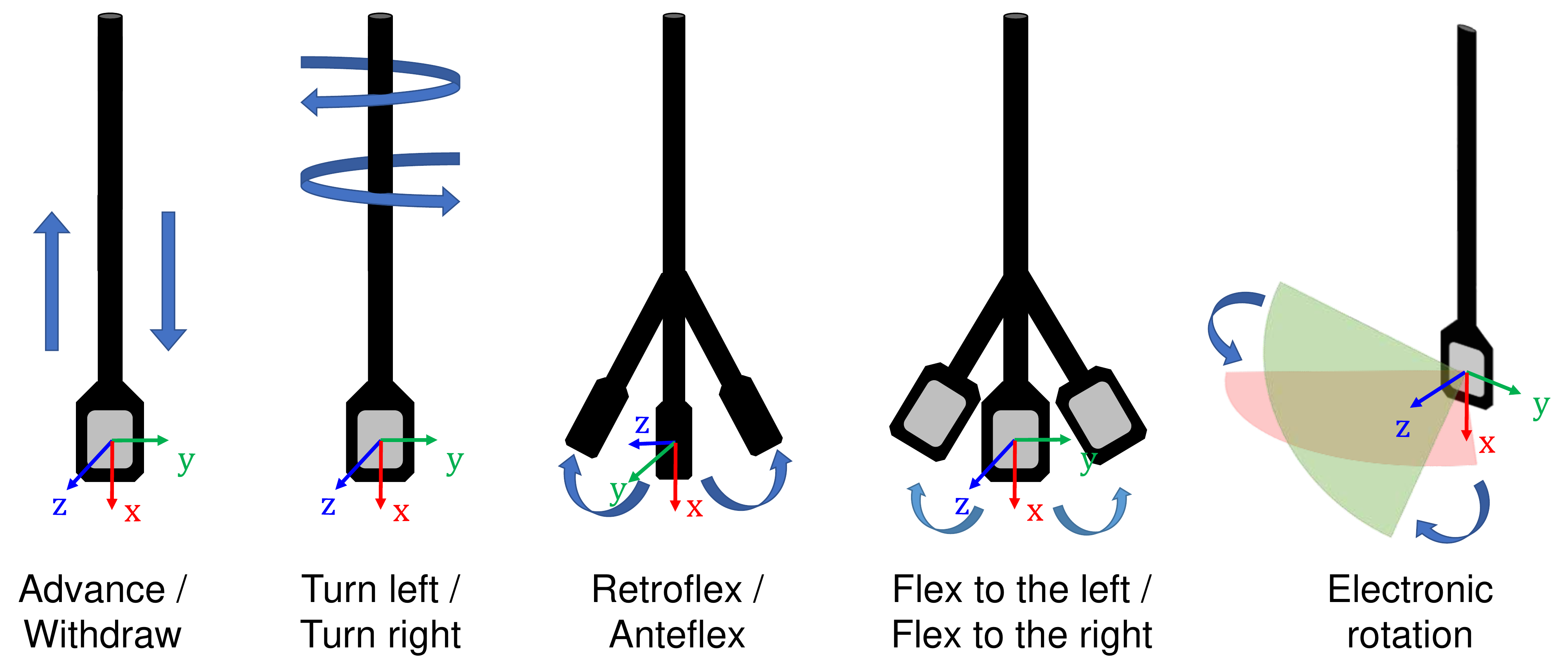}
\caption{The commonly used DOFs of the probe in clinical TEE acquisitions.}
\label{Fig_dofs}
\end{figure}

The commonly used DOFs of the TEE probe in clinical settings are illustrated in Fig. \ref{Fig_dofs}. For simplicity of discussion, we define the probe coordinate frame as follows. The $x$-axis is along the probe's heading direction, the $z$-axis is perpendicular to the transducer array, and the $y$-axis can be obtained by the right-hand rule. Then, the traditional TEE probe manipulation can be described as follows:
\begin{enumerate}
\item ``{Advance / Withdraw}": translation along the $x$-axis; 
\item  ``{Turn left / Turn right}": rotation around the $x$-axis; 
\item  ``{Retroflex / Anteflex}": rotation around the $y$-axis; 
\item  ``{Flex to the left / Flex to the right}": rotation around the $z$-axis; and
\item  ``{Electronic rotation}": rotation of the imaging plane around the $z$-axis from $0^\circ$ to $180^\circ$ 
\end{enumerate}

Since the rotation around the $z$-axis can be adjusted by means of electronic buttons, the system actually only needs to realize 3-DOF control of the probe, i.e., translation along $z$, and rotation around $x,y$ axes of the probe coordinate frame. However, since the position and shape of the patient's esophagus relative to the global world coordinate frame is unknown, the translation of the probe along the $z$-axis can only be achieved if the system can realize 3-DOF position control in a given fixed world coordinate frame. 
By aligning the embedded magnet's axis of magnetization with the $z$-axis of the transducer, we can achieve 2-DOF orientation control in the probe coordinate frame and 3-DOF position control in a fixed world coordinate frame, which ensures the feasibility to provide the required 3 DOFs defined in the probe-centric frame for TEE acquisitions.

\subsection{System Overview}

\begin{figure}[t]
\setlength{\abovecaptionskip}{-0.2cm}
\centering
\includegraphics[scale=1.0,angle=0,width=0.49\textwidth]{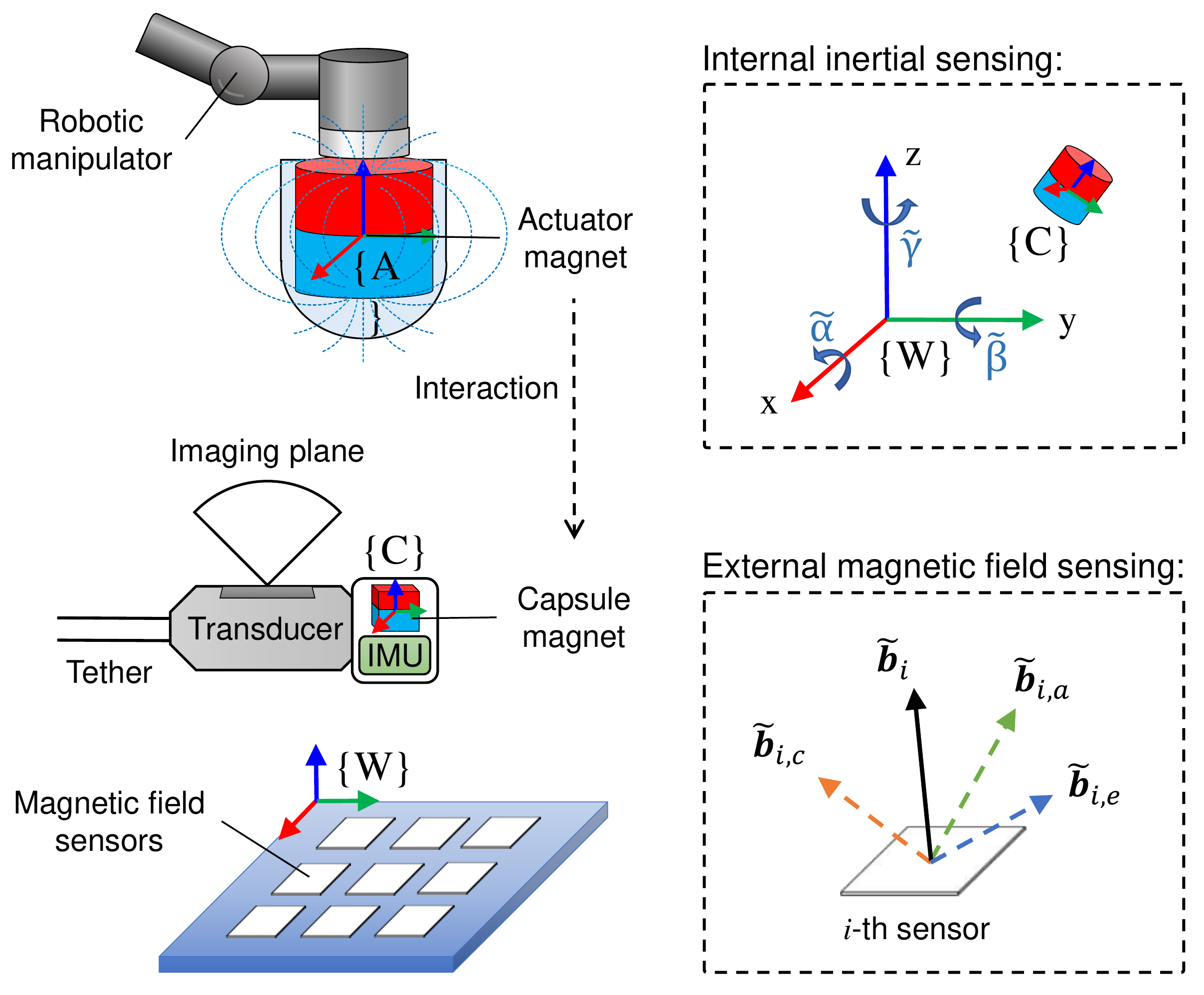}
\caption{Schematic of the MA-TEE system. Combining the measurements from the IMU internal to the probe and the magnetic field sensors external to the probe, the 6-DOF pose of the probe can be estimated in real time and 5-DOF control of the probe can be achieved through interactions between the magnets in the actuator and the capsule.}
\label{Fig_system}
\end{figure}

\begin{figure*}[t]
\centering
\includegraphics[scale=1.0,angle=0,width=0.98\textwidth]{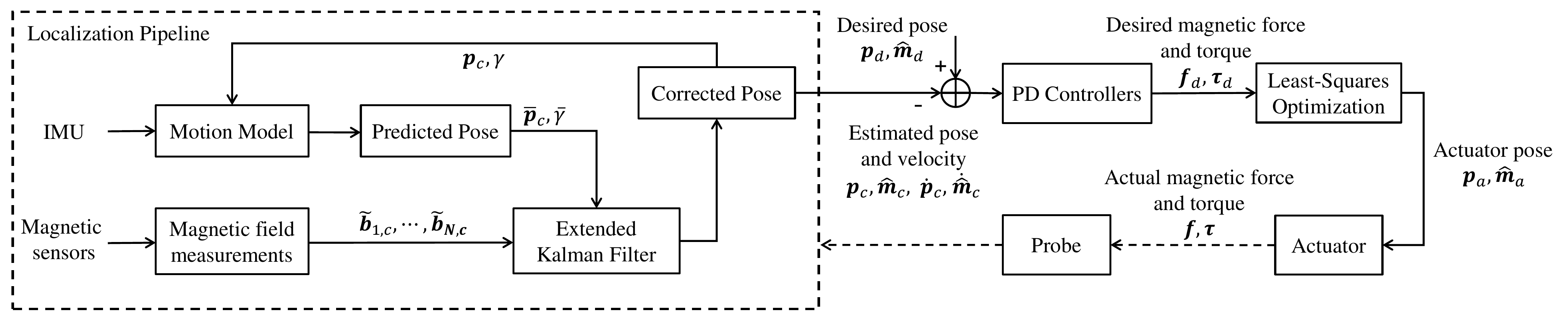}
\caption{The magnetic control workflow of our proposed MA-TEE system. After initialization, closed-loop control of the probe is performed by continuously updating the actuator's pose based on the user-specified desired probe pose and the real-time localization results.}  
\label{Fig_overview_workflow}
\end{figure*}

As illustrated in Fig. \ref{Fig_system}, the MA-TEE probe contains the capsule magnet, an IMU sensor, and an ultrasound transducer with a soft tether, which can be viewed as a tethered capsule robot. We attach a coordinate frame $\{C\}$ to the capsule magnet, and frame $\{A\}$ to the external actuator magnet. We assume that the origins of $\{C\}$ and $\{A\}$ coincide with the center of the magnets, and their positions are $\mathbf{p}_{c}, \mathbf{p}_{a} \in \mathbb{R}^3$, and the $z$-axes of the frames are aligned with the magnetic dipole moments $\mathbf{m}_{c}, \mathbf{m}_{a} \in \mathbb{R}^3$, respectively. 
Note that we assume the capsule magnet frame $\{C\}$ is coincident with the probe coordinate frame as defined in Section~IV-A given that the overall size of the probe is small.
The world coordinate frame $\{W\}$ is attached to the external magnetic sensor array, which has a known and fixed relationship with the robot base. The orientations of the capsule and the actuator w.r.t. the world frame can be represented by matrices $\mathbf{R}_{c}, \mathbf{R}_{a} \in SO(3)$, respectively. Assume that the external magnetic sensor array contains $N$ three-axis magnetic field sensors, whose positions w.r.t. $\{W\}$ are $\mathbf{p}_{i}$, $i = 1,\cdots, N$.

The closed-loop control workflow of our proposed MA-TEE system is illustrated in Fig. \ref{Fig_overview_workflow}. After an initial localization of the probe, the actuator is automatically controlled to move above the probe. Then, the 6-DOF pose of the probe is estimated in real time as shown in the localization pipeline, and 5-DOF magnetic control of the probe is performed to reach the user-specified desired pose for tele-operated TEE acquisitions.

In the following, we first describe the localization algorithm based on external and internal sensor fusion, and then introduce the closed-loop actuation algorithm.

\subsection{Sensor Fusion based Localization}
\subsubsection{Localization Model}

By attaching the capsule magnet and the IMU with the structure as shown in Fig. \ref{Fig_system}, we can estimate the 6D pose of the probe based on the localization model described in \cite{li2022external}. We briefly summarize the mathematics here.
As shown in Fig. \ref{Fig_system}, the orientation of the probe can be represented by the roll, pitch and yaw angles $\alpha, \beta, \gamma$ as $\mathbf{R}_c=\mathbf{R}_z (\gamma)\mathbf{R}_y (\beta)\mathbf{R}_x (\alpha)$, where $\alpha, \beta$ are assumed to be accurately measured by the IMU sensor as $\widetilde{\alpha}, \widetilde{\beta}$. The unknown parameters in the 6D pose of the probe are $\mathbf{p}_c$ and $\gamma$.

Previous work has shown that for an axially magnetized cylindrical magnet with a diameter-to-length ratio of about 1, the percent error of dipole approximation can be reduced to 1\% when the distance is about twice the radius of the minimum bounding sphere of the magnet \cite{petruska2012optimal}.
Since the scale of the probe and the actuator is much smaller than the working distance of the system, we can use the magnetic point-dipole model to estimate the theoretical magnetic field of the probe at the $i$-th sensor's location:

\begin{equation}
\label{F_bc}
\begin{aligned}
\mathbf{b}_{i,c}&=\frac{\mu_{0}\|\mathbf{m}_{c}\|}{4\pi\|\mathbf{r}_i\|^5}\left(3\mathbf{r}_i\mathbf{r}_i^{T}-\|\mathbf{r}_i\|^2\mathbf{I}_{3}\right)\widehat{\mathbf{m}}_{c}, \\
\end{aligned}
\end{equation}

\noindent where $\mu_0$ is the permeability of free space. $\mathbf{r}_i = \mathbf{p}_i - \mathbf{p}_c$ is the vector pointing from the probe to the $i$-th sensor. $\widehat{\mathbf{m}}_{c}=\mathbf{R}_z (\gamma)\mathbf{R}_y (\widetilde{\beta})\mathbf{R}_x (\widetilde{\alpha})[0,0,1]^T$ is a function of $\gamma$.

The actual magnetic field measurement of the probe at the $i$-th sensor $\widetilde{\mathbf{b}}_{i,c}$ can be extracted from the original sensor measurement $\widetilde{\mathbf{b}}_i$ as $\widetilde{\mathbf{b}}_{i,c} = \widetilde{\mathbf{b}}_i - \mathbf{b}_{i,a} - \widetilde{\mathbf{b}}_{i,e}$, where $\widetilde{\mathbf{b}}_{i,e}$ is the environmental magnetic field measurement, and $\mathbf{b}_{i,a}$ is the theoretical value of the actuator's magnetic field approximated using the generalized complete elliptic integral model \cite{derby2010cylindrical}:

\begin{equation} \small
C(k_c,p,c,s) = \int_0^{\frac{\pi}{2}} \frac{(c \cos^2 \varphi + s \sin^2 \varphi)}{(\cos^2 \varphi + p \sin^2 \varphi)\sqrt{\cos^2 \varphi + k_c^2 \sin^2 \varphi}} d\varphi.
\end{equation}

For a longitudinally magnetized cylinder with length $2L$ and radius $R$ with total magnetization $M$, the closed-form magnetic field components expressed in cylindrical coordinates can be obtained as

\begin{align} \small
b_\rho &= \frac{\mu_0 M R}{\pi} \left[  \alpha_+ P_1(k_+) - \alpha_- P_1(k_-)  \right], \\
b_z &= \frac{\mu_0 M R}{\pi (\rho + R)} \left[  \beta_+ P_2(k_+) - \beta_- P_2(k_-)  \right], \\
b_\varphi &= 0,
\end{align}

\noindent where

\begin{align} \small
\alpha_\pm &= \sqrt{(\xi_\pm^2 + (\rho +R)^2)^{-1}}, \\
\beta_\pm &= \xi_\pm \alpha_\pm , \\
k_\pm &= \sqrt{\frac{\xi_\pm^2 + (\rho - R)^2}{\xi_\pm^2 + (\rho + R)^2}} ,\\
\xi_\pm &= z \pm L , \\
\eta &= \frac{\rho - R}{\rho + R}, \\
P_1(k_\pm) &= C(k_\pm,1,1,-1) ,  \\
P_2(k_\pm) &= C(k_\pm,\eta^2,1,\eta).  
\end{align}

Besides, we estimate the environmental magnetic field $\widetilde{\mathbf{b}}_{i,e}$ before introducing the magnets in the workspace and directly use the estimated value during the operation. This is mainly because we assume no ferromagnetic materials are used in the workspace and the main component of the environmental magnetic field is the Earth's magnetic field, which does not change much in a short period of time (e.g., duration of the TEE procedure).

\subsubsection{Initial Localization with Least-Squares Optimization}

\cite{li2022external} uses a non-linear least-squares (LS) optimization method to recover $\mathbf{p}_c, \gamma$ by minimizing the squared distances between the theoretical and measured magnetic fields of the probe from $N$ magnetic sensors $\|\mathbf{b}_{c}- \widetilde{\mathbf{b}}_{c}\|^2$, where $\mathbf{b}_{c} = [\mathbf{b}_{1,c}^T, \cdots,
\mathbf{b}_{N,c}^T]^T$ and $\widetilde{\mathbf{b}}_{c} = [\widetilde{\mathbf{b}}_{1,c}^T, \cdots, \widetilde{\mathbf{b}}_{N,c}^T]^T$.
The method can achieve an average position accuracy of \SI{2.06}{\mm}, \SI{2.86}{\mm}, and \SI{2.81}{\mm} in $x$, $y$, $z$ axes, and an orientation accuracy of \ang{0.19}, \ang{1.02}, and \ang{1.01} in $x$, $y$, $z$ axes, in a \SI[parse-numbers=false]{0.5 $\times$ 0.5 $\times$ 0.2}{\cubic\m} workspace \cite{li2022external}.
In this paper, we use this method for initial localization as it does not rely on previous pose estimation results.

\subsubsection{Real-time Tracking with Extended Kalman Filter}
In order to better deal with the noisy sensor readings and take advantage of the historical information, in this work, we improve upon our previous work \cite{li2022external} by applying an Extended Kalman Filter (EKF) to combine  IMU-based motion prediction with the magnetic field measurements for real-time tracking of the probe.
Let $\mathbf{x}_t = \begin{bmatrix}\mathbf{p}_{c}^{(t)} \\ \gamma^{(t)} \end{bmatrix} \in \mathbb{R}^4$ denote the state and $\mathbf{z}_t=\widetilde{\mathbf{b}}_{c}^{(t)} \in \mathbb{R}^{3N}$ denote the magnetic field measurements at time step $t$. Consider the following system model:

\begin{align}
\label{F_system_model}
\, \mathbf{x}_t &= f(\mathbf{x}_{t-1}, \mathbf{u}_{t-1}) + \mathbf{w}_t,
\\
\, \mathbf{z}_t &= h(\mathbf{x}_{t}) + \mathbf{n}_t,
\end{align}

\noindent where $f(\cdot)$ is the motion model, $\mathbf{u}_{t-1}=\begin{bmatrix} \mathbf{a}^{(t-1)} \\ \boldsymbol{\omega}^{(t-1)} \end{bmatrix} \in \mathbb{R}^6$ is the acceleration and rotational velocity of the probe measured by the IMU at the previous time step, $\mathbf{w}_t \sim \mathcal{N}(\mathbf{0}, \mathbf{Q}_t)$ is the motion model noise with zero-mean and covariances $\mathbf{Q}_t$. $h(\cdot)$ is the measurement model $\mathbf{b}_{c}$, and $\mathbf{n}_t \sim \mathcal{N}(\mathbf{0}, \mathbf{N}_t)$ is the measurement model noise with zero-mean and covariances $\mathbf{N}_t$.

Given the previous probe pose $\mathbf{p}_{c}^{(t-1)},\mathbf{R}_{c}^{(t-1)}$, the probe pose at the time step $t$ can be predicted as

\begin{align}
 \bar{\mathbf{p}}_c^{(t)} &= \mathbf{p}_c^{(t-1)} + \mathbf{v}_c^{(t-1)}\Delta t + \frac{1}{2}(\mathbf{R}_c^{(t-1)}\mathbf{a}^{(t-1)}+\mathbf{g}) \Delta t^2 \label{p_pred},
\\
 \bar{\mathbf{v}}_c^{(t)} &= \mathbf{v}_c^{(t-1)} + (\mathbf{R}_c^{(t-1)}\mathbf{a}^{(t-1)}+\mathbf{g}) \Delta t \label{v_pred},
\\
\bar{\mathbf{R}}_c^{(t)} &= \mathbf{R}_c^{(t-1)}\mathbf{R}\{\boldsymbol{\omega}^{(t-1)}\Delta t\}, 
\end{align}

\noindent where $\mathbf{v}_c^{(t)}$ is the velocity of the probe at time step $t$, $\mathbf{g}$ is the acceleration of gravity, $\mathbf{R}\{\boldsymbol{\omega}^{(t-1)}\Delta t\}$ is the rotation matrix equivalent to the axis-angle representation $\boldsymbol{\omega}^{(t-1)}\Delta t$. By changing the orientation representation as Euler angles, we can obtain the predicted state $\bar{\mathbf{x}}_{t} = \begin{bmatrix}\bar{\mathbf{p}}_{c}^{(t)} \\ \bar{\gamma}^{(t)} \end{bmatrix}$ at time step $t$. 
Since the motion of the probe tip in the esophagus is mainly in a stick-slip fashion, we assume zero-velocity at the previous step, i.e.,  $\mathbf{v}_c^{(t-1)}=\mathbf{0}$ in our implementation.

The covariance matrix of the state is predicted as

\begin{equation}
\label{F_covariances}
\bar{\mathbf{P}}_{t} = \mathbf{F} \mathbf{P}_{t-1} \mathbf{F}^T + \mathbf{Q}_t,
\end{equation}

\noindent where $\mathbf{F}=\frac{\partial f}{\partial \mathbf{x}_{t-1}}|_{ \mathbf{x}_{t-1}}$ is the Jacobian of $f(\cdot)$. Since the pose change in the period $\Delta t$ is small, we approximate the motion model Jacobian as $\mathbf{F} \doteq \mathbf{I}_4$, and $\mathbf{Q}_t$ is empirically chosen as $\mathbf{Q}_t = \text{diag}(0.01,0.01,0.01,30)$ to indicate an uncertainty of \SI{0.01}{\m} in position and \ang{30} in yaw estimation.

In the correction step, the Kalman gain is computed as

\begin{equation}
\label{F_kalman_gain}
\mathbf{K} = \bar{\mathbf{P}}_{t} \mathbf{H}^T (\mathbf{H} \bar{\mathbf{P}}_{t} \mathbf{H}^T + \mathbf{N}_t)^{-1},
\end{equation}

\noindent where $\mathbf{H}$ is the Jacobian of $h(\mathbf{x}_{t})= [\mathbf{b}_{1,c}^T, \cdots, \mathbf{b}_{N,c}^T]^T$, which can be calculated as

\begin{equation}
\mathbf{H} =\left. \frac{\partial h}{\partial \mathbf{x}_{t}}\right\vert_{ \bar{\mathbf{x}}_{t}} = [\mathbf{J}_1^T, \cdots, \mathbf{J}_N^T],
\end{equation}
\begin{equation}
\mathbf{J}_i = \left[ \frac{\partial \mathbf{b}_{i,c}}{\partial \mathbf{p}_c} \ \  \frac{\partial \mathbf{b}_{i,c}}{\partial \gamma} \right]\bigg|_{\bar{\mathbf{p}}_c^{(t)}, \bar{\gamma}^{(t)}}, i \in \{1, \cdots, N\}, \\
\end{equation}

\noindent where

\begin{small} 
\begin{align}
&\frac{\partial \mathbf{b}_{i,c}}{\partial \mathbf{p}_c} = -\frac{3 \mu_{0}\|\mathbf{m}_c\|}{4\pi\|\mathbf{r}_i\|^5} 
\Big[ \Big( \mathbf{r}_i^T \widehat{\mathbf{m}}_c \Big) 
\Big( \mathbf{I}_3-\frac{5  \mathbf{r}_i \mathbf{r}_i^T}{\|\mathbf{r}_i\|^2} \Big)
+ \widehat{\mathbf{m}}_c \mathbf{r}_i^T  + \mathbf{r}_i \widehat{\mathbf{m}}_c ^T \Big], \\
& \frac{\partial \mathbf{b}_{i,c}}{\partial \gamma}  =
\frac{\mu_{0} \|\mathbf{m}_c\|}{4\pi\|\mathbf{r}_i\|^5}\left(3\mathbf{r}_i\mathbf{r}_i^{T}-\|\mathbf{r}_i\|^2\mathbf{I}_{3}\right)\frac{\partial \widehat{\mathbf{m}}_c}{\partial \gamma}, \\
&\frac{\partial \widehat{\mathbf{m}}_c}{\partial \gamma} = 
\left[ \begin{matrix} \sin \widetilde{\alpha} \cos {\gamma} - \cos \widetilde{\alpha} \sin \widetilde{\beta} \sin {\gamma} \\ 
\sin \widetilde{\alpha} \sin {\gamma} + \cos \widetilde{\alpha} \sin \widetilde{\beta} \cos {\gamma} \\ 0 \end{matrix} \right].
\end{align}
\end{small}

Then, the state of the probe is updated as

\begin{equation}
\label{F_x_update}
\mathbf{x}_{t} = \bar{\mathbf{x}}_{t} + \mathbf{K}\big(\mathbf{z_t}-h(\bar{\mathbf{x}}_{t})\big).
\end{equation}

The covariance matrix is updated using the symmetric and positive \textit{Joseph} form \cite{joseph}

\begin{equation}
\label{F_covariances_update}
\mathbf{P}_{t} = (\mathbf{I}_4 - \mathbf{K}\mathbf{H})\bar{\mathbf{P}}_{t}(\mathbf{I}_4 - \mathbf{K}\mathbf{H})^T + \mathbf{K}\mathbf{N}_t\mathbf{K}^T,
\end{equation}
\noindent where the measurement noise covariance matrix is set as $\mathbf{N}_t=\sigma^2 \cdot \mathbf{I}_{3N}$, $\sigma = \SI{1e-5}{\tesla}$ based on our experimental results.

\begin{algorithm}[t] \small
\caption{Sensor Fusion based Localization}
\label{Alg_Localization}
\KwIn{{positions of $N$ magnetic field sensors $\mathbf{p}_{1}, \cdots, \mathbf{p}_{N}$, 
environmental magnetic fields $\widetilde{\mathbf{b}}_{1,e} \cdots, \widetilde{\mathbf{b}}_{N,e}$, 
IMU measurements $\mathbf{a}^{(t)}, \boldsymbol{\omega}^{(t)}, \widetilde{\alpha}, \widetilde{\beta}$,
actuator's pose $\mathbf{p}_{a}^{(t)},\mathbf{R}_{a}^{(t)}$,
real-time magnetic field measurements $\widetilde{\mathbf{b}}_1^{(t)}, \cdots, \widetilde{\mathbf{b}}_N^{(t)}$,
$t=0, 1, 2,\cdots$}}
\KwOut{6-DOF pose of the probe $\mathbf{p}_{c}^{(t)},\mathbf{R}_{c}^{(t)}$, $t=0, 1, 2,\cdots$}
Obtain the initial theoretical and measured magnetic field of the probe at each sensor $\mathbf{b}_{i,c}^{(0)}$, $\widetilde{\mathbf{b}}_{i,c}^{(0)}$, $i=1, \cdots, N$\;
Perform initial localization to solve $\mathbf{p}_{c}^{(0)},\gamma^{(0)}$\;
Obtain the initial orientation of the probe $\mathbf{R}_{c}^{(0)}$ based on $\widetilde{\alpha}^{(0)}, \widetilde{\beta}^{(0)},\gamma^{(0)}$\;
\For{$t=1,2,\cdots$}
{
Predict the probe pose $\mathbf{p}_{c}^{(t)},\mathbf{R}_{c}^{(t)}$ based on the last pose estimate $\mathbf{p}_{c}^{(t-1)},\mathbf{R}_{c}^{(t-1)}$ and IMU measurements $\mathbf{a}^{(t-1)}, \boldsymbol{\omega}^{(t-1)}$ \;
Obtain the theoretical and measured magnetic field of the probe at each sensor $\mathbf{b}_{i,c}^{(t)}$, $\widetilde{\mathbf{b}}_{i,c}^{(t)}$, $i=1, \cdots, N$\;
Update the pose estimate $\mathbf{p}_{c}^{(t)},\gamma^{(t)}$ with EKF\;
Update the orientation of the probe $\mathbf{R}_{c}^{(t)}$ based on $\widetilde{\alpha}^{(t)}, \widetilde{\beta}^{(t)},\gamma^{(t)}$\;
}
\Return {6-DOF pose of the probe} $\mathbf{p}_{c}^{(t)},\mathbf{R}_{c}^{(t)}$, $t=1,2,\cdots$\;
\end{algorithm}

The localization algorithm is summarized in Algorithm \ref{Alg_Localization}.

\subsection{Closed-Loop Magnetic Actuation}

\subsubsection{Dynamic Model of the Probe}

Similar to most existing closed-loop magnetic control systems, our control approach is developed by modelling the magnetic fields of the actuator and the probe using the magnetic point-dipole model. Given the poses of the probe and the actuator magnet, the magnetic force and torque applied to the probe can be calculated as

\begin{align}
\label{F_f}
\boldsymbol{f}& = \nabla \,(\mathbf{m}_c \cdot \mathbf{b}), 
\\
\boldsymbol{\tau}& =  \mathbf{m}_c \times \mathbf{b},
\end{align}

\noindent where $\nabla$ is the gradient operator, $\mathbf{b}$ is the magnetic field of the actuator at the probe's position:

\begin{equation}
\label{F_b}
\mathbf{b}=\frac{\mu_{0}\|\mathbf{m}_a\|}{4\pi\|\mathbf{p}\|^5}\left(3\mathbf{p}\mathbf{p}^{T}-\|\mathbf{p}\|^2\mathbf{I}_{3}\right)\widehat{\mathbf{m}}_a, \\
\end{equation}

\noindent where $\mathbf{p} = \mathbf{p}_{c} - \mathbf{p}_{a}$ is the vector pointing from $\mathbf{p}_{a}$ to $\mathbf{p}_{c}$.

Different from previous work that focused on the magnetic actuation of a capsule in fluid \cite{mahoney2016five} or in an air-filled colon with little contact with the colon walls \cite{pittiglio2019magnetic}, in our target application, the esophagus has a small diameter of about \SI{20}{\mm} and is collapsed between swallows \cite{kuo2006esophagus}, and the TEE acquisition requires the probe tip to continuously contact the esophageal wall to ensure acoustic coupling. Therefore, the friction between the probe tip and the esophagus should be modeled in our control method. 
Based on the experimental data in \cite{lin2017friction}, in this work, we estimate the friction force on the probe-tissue interface as a constant resistance force in the opposite direction of velocity as

\begin{equation}
\label{F_res}
\boldsymbol{f}_{\textit{res}}= - f_{\textit{res}} \, \frac{\dot{\mathbf{p}}_c}{\| \dot{\mathbf{p}}_c \|}.  \  \\
\end{equation}

Similar to \cite{pittiglio2019magnetic}, we consider the impact of the tether behind the probe as an unmodelled disturbance, which can actually serve as a stabilizing damper to improve the stability of the system.  Then, the total force applied to the probe is estimated as the sum of the magnetic force, the force of gravity on the probe, and the resistance forces in the esophagus. The total torque is assumed as the magnetic torque only. Therefore, the dynamic model of the probe can be represented as

\begin{equation}
\label{F_dynamics}
\begin{cases}
\, m_c \ddot{\mathbf{p}}_c= \boldsymbol{f} + \boldsymbol{f}_g + \boldsymbol{f}_{\textit{res}}\\
\ \mathbf{I}_c \, \boldsymbol{\alpha} = \,\boldsymbol{\tau}
\end{cases},
\end{equation}

\noindent where $m_c$ and $\mathbf{I}_c$ are the mass and inertia tensor of the probe, $\ddot{\mathbf{p}}_c$ and $\boldsymbol{\alpha}$ are the position and angular accelerations of the probe, and $\boldsymbol{f}_g$ is the force of gravity on the probe.

\subsubsection{Closed-Loop Actuation Algorithm}

\begin{algorithm}[t] \small 
\caption{Closed-Loop Magnetic Actuation}
\label{Alg_actuation}
\KwIn{Estimated probe pose $\mathbf{p}_{c}^{(t)}, \widehat{\mathbf{m}}_c^{(t)}$, desired probe pose $\mathbf{p}_d^{(t)}, \widehat{\mathbf{m}}_d^{(t)}$, $t=1,2,\cdots$}
\KwOut{5-DOF pose of the actuator $\mathbf{p}_{a}^{(t)}$, $\widehat{\pmb{m}}_{a}^{(t)}$, $t=1,2,\cdots$}
\For{$t=1, 2, \cdots$}
{
Estimate the probe's velocity $\dot{\mathbf{p}}_{c}^{(t)}, \dot{\widehat{\mathbf{m}}}_c^{(t)}$\;
Calculate the pose and velocity errors\;
Obtain the desired magnetic force and torque $\boldsymbol{f}_{d}, \boldsymbol{\tau}_d$ with PD controllers\;
Update the actuator pose $\mathbf{p}_{a}^{(t)}$, $\widehat{\pmb{m}}_{a}^{(t)}$ using (\ref{F_actuator_update})\;
}
\Return{5-DOF pose of the actuator} $\mathbf{p}_{a}^{(t)}$, $\widehat{\pmb{m}}_{a}^{(t)}$, $t=1,2,\cdots$\;
\end{algorithm}

The goal of actuation is to reach the desired 5-DOF pose of the TEE probe tip $\mathbf{p}_d, \widehat{\mathbf{m}}_d$ for ultrasound image acquisition. To this end, two separate PD controllers are developed to ensure the position and orientation errors between the current and desired poses remain small. The control scheme is illustrated in Fig. \ref{Fig_overview_workflow}.
Given the pose estimation $\mathbf{p}_{c}, \widehat{\mathbf{m}}_c$ from the localization algorithm, the velocity of the probe $\dot{\mathbf{p}}_{c}, \dot{\widehat{\mathbf{m}}}_c$ can be estimated by calculating the simple moving average of pose differences between consecutive steps over a short time period. 
The position error between the current pose and the desired pose is $\mathbf{e}_p = \mathbf{p}_d - \mathbf{p}_c$, and the velocity error is $\dot{\mathbf{e}}_p = - \dot{\mathbf{p}}_{c}$. The orientation error between the current and desired magnetic dipole moments is calculated as 
$\mathbf{e}_o= \widehat{\mathbf{m}}_c \times \widehat{\mathbf{m}}_d$, and the rotational velocity error is calculated as $\dot{\mathbf{e}}_o= \dot{\widehat{\mathbf{m}}}_c \times \widehat{\mathbf{m}}_d$.
Then, the desired force and torque applied to the probe are given by:

\begin{equation}
\label{F_fd}
\begin{cases}
\boldsymbol{f}_{d} = \mathbf{K}_p \mathbf{e}_p + \mathbf{K}_d \dot{\mathbf{e}_p} - \boldsymbol{f}_g - \boldsymbol{f}_{\textit{res}} \\
\boldsymbol{\tau}_d =  \mathbf{K}_{po} \mathbf{e}_{o} + \mathbf{K}_{do} \dot{\mathbf{e}_o}  \\
\end{cases}.
\end{equation}

Since the theoretical magnetic force and torque can be calculated using (\ref{F_f}) as a function of the actuator's pose $\mathbf{p}_{a}, \widehat{\mathbf{m}}_a$, we can update the desired actuator pose by solving the constrained non-linear least-squares problem as follows:

\begin{equation} \small
\label{F_actuator_update}
\begin{aligned}
\mathbf{p}_{a}^*, \mathbf{m}_a^* = \mathop{\arg\min}_{\mathbf{p}_{a}, \mathbf{m}_a} \quad & \bigg\| \left[ \begin{matrix}\boldsymbol{f}_{d}\\\boldsymbol{\tau}_d\end{matrix} \right] - \left[ \begin{matrix}\boldsymbol{f}(\mathbf{p}_{a}, {\mathbf{m}}_a)\\\boldsymbol{\tau}(\mathbf{p}_{a}, {\mathbf{m}}_a) \end{matrix} \right] \bigg\|^2,\\
\textrm{subject to} \quad
& \mathbf{p}_{a} \in \mathbb{W}, \\
 & \mathbf{m}_a \in \mathbb{R}^3, \\
\end{aligned}
\end{equation}

\noindent where $\mathbb{W} \subseteq \mathbb{R}^3$ is used to limit the movement of the actuator for safety reasons. The unit magnetic dipole moment of the actuator $\widehat{\mathbf{m}}_a$ is then updated by normalizing $\mathbf{m}_a^*$. 

The closed-loop magnetic actuation algorithm is summarized in Algorithm \ref{Alg_actuation}. 

\begin{figure*}[tb]
\centering
\includegraphics[scale=1.0,angle=0,width=0.9\textwidth]{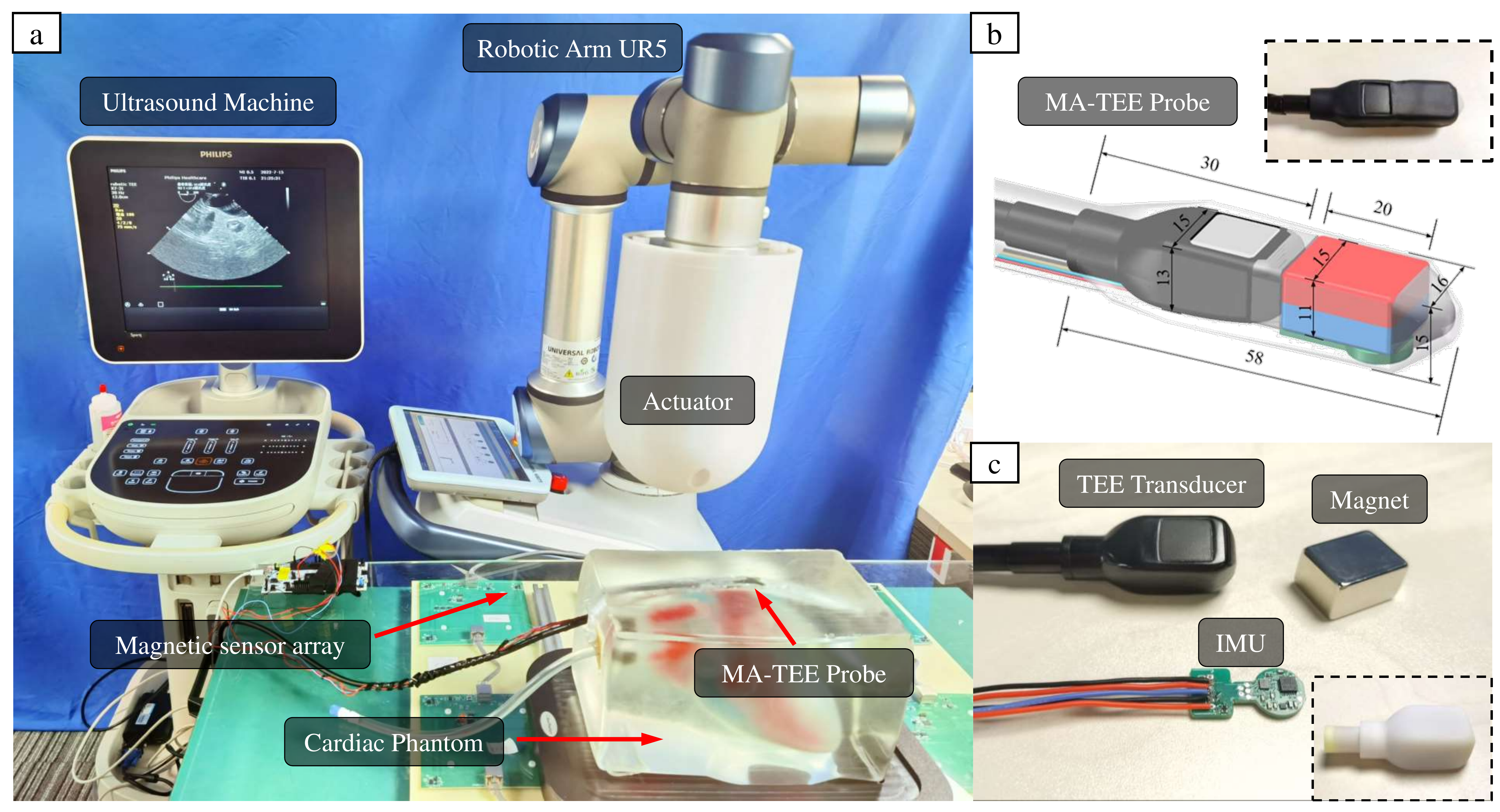}
\caption{(a) shows the experimental setup of the proposed MA-TEE system. The actuator magnet is rigidly mounted at the end-effector of a UR5 robotic arm for closed-loop control of the probe in the simulated esophagus in a cardiac phantom. The workspace of the system is above the external sensor array placed on a horizontal platform. 
(b) is a closer look at the MA-TEE probe prototype. All dimensions are in millimeters (\si{\mm}) and (c) illustrates its components, including a TEE transducer, a cuboid permanent magnet and an IMU. The black dashed box in the lower right corner shows a dummy transducer fabricated for use in our quantitative experiments.}
\label{Fig_system_setup}
\end{figure*}

\section{Experiments}
\subsection{System Setup} \label{section_system_setup}

An overview of our real-world system setup is shown in Fig. \ref{Fig_system_setup}(a). The actuator is a cylindrical NdFeB magnet (N45 grade, axially magnetized, diameter and length: \SI{90}{\mm}) with a 3D-printed shell, which is attached to the end-effector of a 6-DOF robotic arm (UR5, Universal Robots). The distance from the center of the magnet to the robot flange is set to \SI{100}{\mm} to ensure normal operation of the motors in the robotic arm.
The external magnetic field sensor array on the horizontal platform is composed of $6 \times 6$ three-axis high-precision magnetic field sensors (IIS2MDC, STMicroelectronics) with a spacing of \SI{120}{\mm}, which are arranged on $9$ circuit boards with 4 sensors per board \cite{li2022external}. The magnetic sensors are connected to a desktop via cables, and the output rate is \SI{100}{\hertz}. Ferromagnetic materials were removed from the workspace to minimize magnetic interference and ensure safety of the procedure. The transformation between the robot base and the external sensor array was measured before the experiments and fixed during the experiments. The localization and closed-loop control algorithms were implemented on the desktop using Python. 

The MA-TEE probe prototype was modified from a standard adult TEE probe (X7-2t, Philips Healthcare) as follows. First, the flexible gastroscope in the traditional TEE probe was replaced with a soft tether that contains only wires to transmit ultrasound and IMU data. Then, the probe tip was attached to a cuboid NdFeB magnet (N52 grade, $20 \times 15 \times \SI{11}{\cubic \mm}$) and an IMU sensor (ICM-20948, InvenSense, TDK) using 3D-printed connectors, and wrapped with thermoplastic elastomers (TPE), as shown in Fig. \ref{Fig_system_setup}(b). The overall size of the MA-TEE probe is $58 \times 16 \times \SI{15}{\cubic \mm}$, and the soft tether has a diameter of \SI{6}{\mm}. In comparison, a standard X7-2t TEE probe has a size of $38 \times 17 \times \SI{13.5}{\cubic \mm}$, and the shaft has a diameter of \SI{10}{\mm} \cite{pushparajah2012}. The MA-TEE probe was connected to a medical ultrasound machine (Sparq, Philips Healthcare) and the ultrasound images were streamed to the desktop via a data acquisition card. The data from the IMU is transmitted to the desktop through wires at \SI{100}{\hertz}. 
We also fabricated a mockup MA-TEE probe to quantitatively evaluate the localization and closed-loop control methods as described in Section~\ref{section_exp_localization} and \ref{section_exp_actuation}, in order to avoid damaging the real ultrasound transducer. The dummy probe contains the same internal magnet and IMU as the MA-TEE probe prototype, and a 3D-printed dummy transducer with the same shape and mass as the real transducer, as shown in the lower right corner of Fig. \ref{Fig_system_setup}(b).

\begin{figure*}[tbh]
\setlength{\abovecaptionskip}{-0.1cm}
\centering
\includegraphics[scale=1.0,angle=0,width=0.8\textwidth]{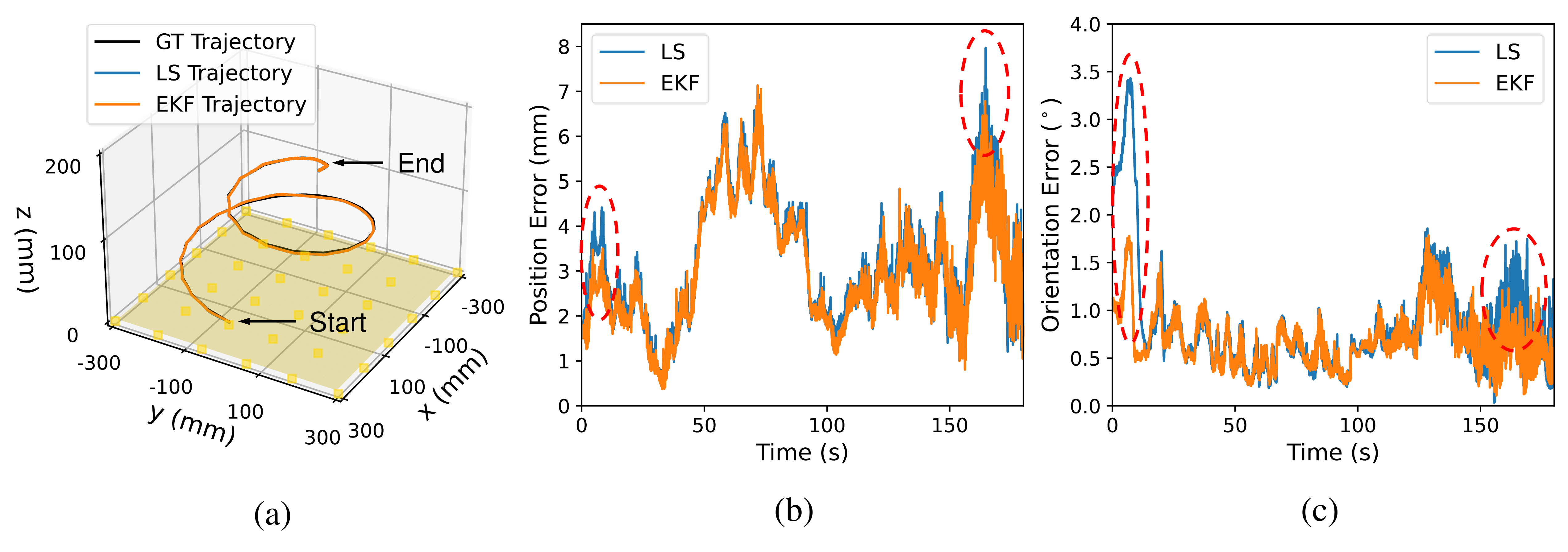}
\caption{(a) shows the moving trajectory of the probe estimated by the LS and  EKF methods compared with the ground truth (GT) when the probe is moving along a 3D spiral trajectory. (b) and (c) show the position and orientation errors during the movement. The red dashed circles highlight some locations where the large errors of the LS method are smoothed by the EKF method.}
\label{Fig_exp_localization_spiral}
\end{figure*}

\begin{figure}[t]
\setlength{\abovecaptionskip}{-0.1cm}
\centering
\includegraphics[scale=1.0,angle=0,width=0.4\textwidth]{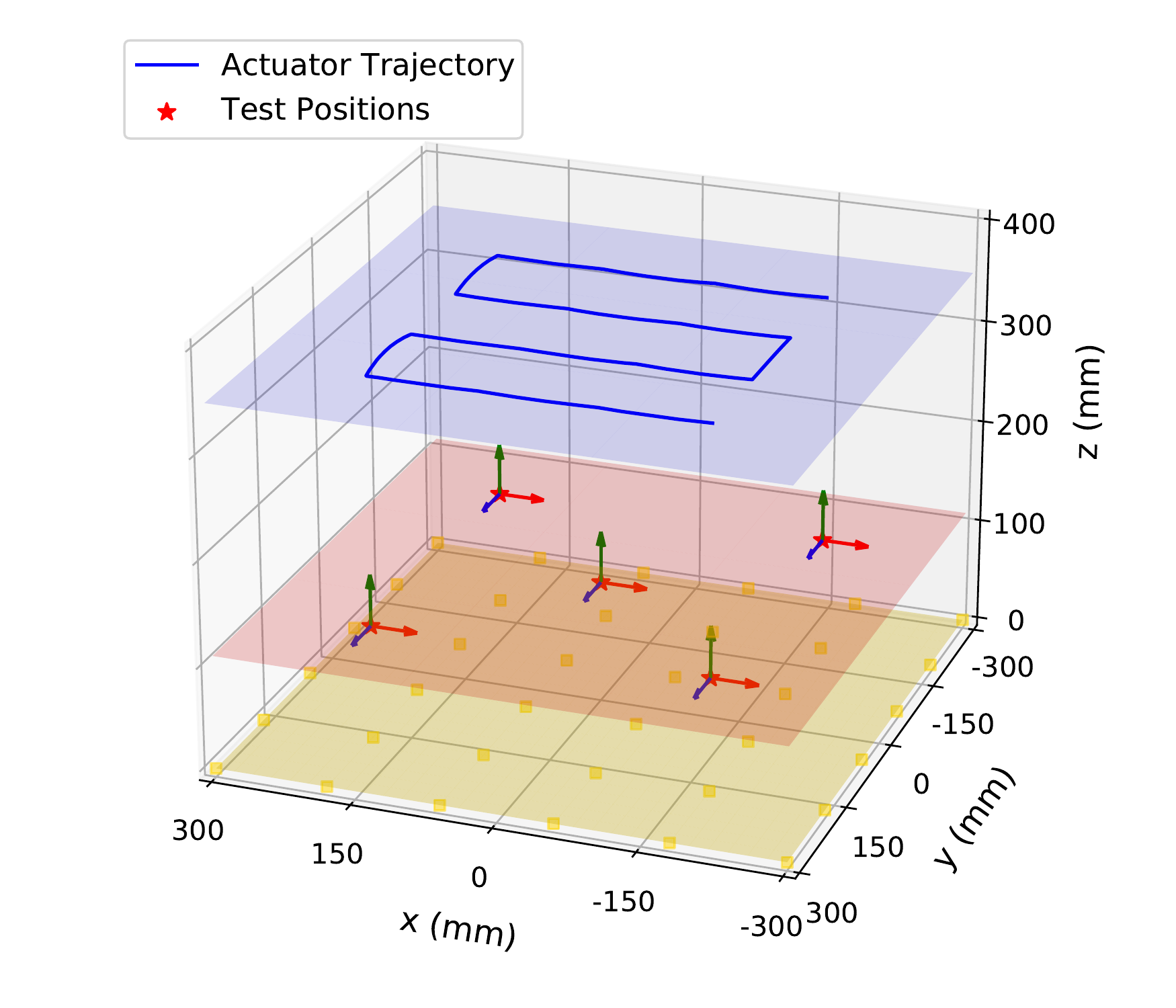}
\caption{Illustration of the localization experiments of the probe in the presence of a dynamic actuator. The probe is mounted at five test positions (\SI{-180}{\mm}, \SI{-180}{\mm}), (\SI{-180}{\mm}, \SI{180}{\mm}) , (\SI{0}{\mm}, \SI{0}{\mm}), (\SI{180}{\mm}, \SI{-180}{\mm}), and (\SI{180}{\mm}, \SI{180}{\mm}) on each horizontal plane (red) at four different heights from \SI{50}{\mm} to \SI{200}{\mm} when the actuator is moving along an ``S"-shaped trajectory (blue) above the plane. }
\label{Fig_exp_localization_quantitative}
\vspace{-0.3cm}
\end{figure}

\begin{table*}[t] \renewcommand\arraystretch{1.2} \small 
\setlength{\abovecaptionskip}{-0.02cm}
\centering
\caption{Localization Accuracy in Position and Orientation (Mean $\pm$ Standard Deviation) of the MA-TEE Probe Rigidly Mounted at Five Points with Four Different Heights When the Actuator is Moving Over It}
\resizebox{0.9\textwidth}{14 mm}{
\begin{tabular}{m{1.5cm}<{\centering}m{2.2cm}<{\centering}m{2.2cm}<{\centering}m{2.2cm}<{\centering}m{2.2cm}<{\centering}m{2.2cm}<{\centering}m{2.2cm}<{\centering}}
\toprule 
\multirow{2}{*}{\tabincell{c}{Height \\(\si{\mm})}}  & \multicolumn{3}{c}{Position error (\si{\mm})}&	\multicolumn{3}{c}{Orientation error (\si{\degree})} \\
\cline{2-7}
& $x$& $y$ & $z$ &$x$ &$y$ & $z$ \\
\midrule
50	&-2.23 $\pm$ 6.24	&1.86 $\pm$ 6.99 & \ 4.69 $\pm$ 8.50	&5.71 $\pm$ 7.57	&0.23 $\pm$ 0.20& 5.72 $\pm$ 7.56 \\
100	&-2.11 $\pm$ 3.78	&1.75 $\pm$ 1.17	&-0.18 $\pm$ 1.15	&1.05 $\pm$ 0.81		&0.26 $\pm$ 0.31	&1.12 $\pm$ 0.78 \\
150	&-1.39 $\pm$ 6.18	&2.06 $\pm$ 3.73	&-0.48 $\pm$ 3.31	&3.11 $\pm$ 2.10
	&	0.34 $\pm$ 0.42	&3.14 $\pm$ 2.09 \\
200	&-2.42 $\pm$ 8.10	&3.97 $\pm$ 5.16&
		-2.23 $\pm$ 8.12	&	4.11 $\pm$ 3.35	&0.29 $\pm$ 0.34	&4.13 $\pm$ 3.34 \\
Average & -2.03 $\pm$ 6.28 & 2.47 $\pm$ 4.61 & \ 0.06 $\pm$ 6.34  & 3.29 $\pm$ 4.19 & 0.28 $\pm$ 0.34 & 3.33 $\pm$ 4.18 \\
\bottomrule
\end{tabular}}
\label{T_localization}
\vspace{-0.3cm}
\end{table*}

\subsection{Evaluation  of the Localization Method} \label{section_exp_localization}

First, we experimentally evaluated the performance of the proposed EKF-based method and the LS-based method in \cite{li2022external} for real-time localization of the probe. In this set of experiments, the mockup MA-TEE probe was rigidly connected to the flange of the robotic arm using a 3D-printed connector and was controlled to move along a 3D spiral trajectory in the \SI[parse-numbers=false]{0.5 $\times$ 0.5 $\times$ 0.2}{\cubic\m} workspace, as shown in Fig. \ref{Fig_exp_localization_spiral}(a). The moving speed of the probe was set to \SI{10}{mm/s}. The ground-truth pose of the probe was provided by the robotic arm. The pose estimation errors were calculated as the Euclidean distance between the positions and the minimum angle required to rotate from the estimated orientation to the true orientation. As shown in Fig. \ref{Fig_exp_localization_spiral}, the EKF method achieves a slightly better localization accuracy of $3.06 \pm \SI{1.33}{mm}$ and $0.69 \pm \ang{0.29}$ compared with the LS method ($3.21 \pm \SI{1.36}{mm}$ and $0.83 \pm \ang{0.57}$). It can be seen from Fig. \ref{Fig_exp_localization_spiral}(b-c) that large errors occasionally occur for the LS method as it only uses the current sensor measurements for localization. In contrast, the EKF method can take advantage of the previous pose estimation by fusing the IMU and magnetic field sensing data, which demonstrated a better capability to deal with noisy sensor measurements and provides a smoother trajectory. In addition, the EKF method provides a higher localization update frequency of \SI[parse-numbers=false]{80 \sim 90}{\hertz}, which is faster than the LS method (\SI[parse-numbers=false]{50\sim 60}{\hertz}).

Furthermore, in order to validate the compatibility of the EKF-based localization method with magnetic actuation, we conducted a set of localization experiments at a total of 20 positions in the 3D workspace above the sensor array in the presence of a dynamic actuator. 4 horizontal planes were selected with an increasing height from \SI{50}{mm} to \SI{200}{mm}, and 5 test points were selected on each plane, as shown in Fig. \ref{Fig_exp_localization_quantitative}. At each test position, the probe was rigidly mounted, and the actuator was commanded to move along an ``S"-shaped trajectory on a horizontal plane at a height of \SI{250}{mm} above the probe. The real-time localization results at each point were recorded during the movement of the actuator, and the localization accuracy in all DOFs is reported in Table \ref{T_localization}. In all trials, the horizontal platform was firmly and rigidly connected to the sensor array using 3D-printed connectors with known size. The rigid connection was used to prevent motion of the probe in the world coordinate frame and provide the ground truth of the probe pose. 

It was found that the best localization performance was achieved when the probe was placed at medium heights (e.g., between \SI{100}{mm} and \SI{150}{mm}) above the sensor array. When the probe is farther away from the sensor array, the signal-to-noise ratio at the magnetic field sensors will be lower, which would result in a decreased localization accuracy. On the other hand, when the probe was too close to the sensor array, the actuator would also be very close to the sensor array to realize effective actuation of the probe, and the inaccuracy of the magnetic field model of the actuator would have a greater impact on the localization performance. 
The average position errors in the \SI[parse-numbers=false]{0.5 $\times$ 0.5 $\times$ 0.2}{\cubic\m} workspace in the presence of a dynamic actuator are \SI{-2.03}{\mm}, \SI{2.47}{\mm}, and \SI{0.06}{\mm} in $x$, $y$, and $z$ axes, respectively, and the average orientation errors are \ang{3.29}, \ang{0.28}, and \ang{3.33} in $x$, $y$, and $z$ axes, respectively. Given the results, we can conclude that the localization system can satisfy the accuracy requirements and be compatible with magnetic actuation.

\begin{figure*}[t]
\setlength{\abovecaptionskip}{-0.1cm}
\centering
\includegraphics[scale=1.0,angle=0,width=0.99\textwidth]{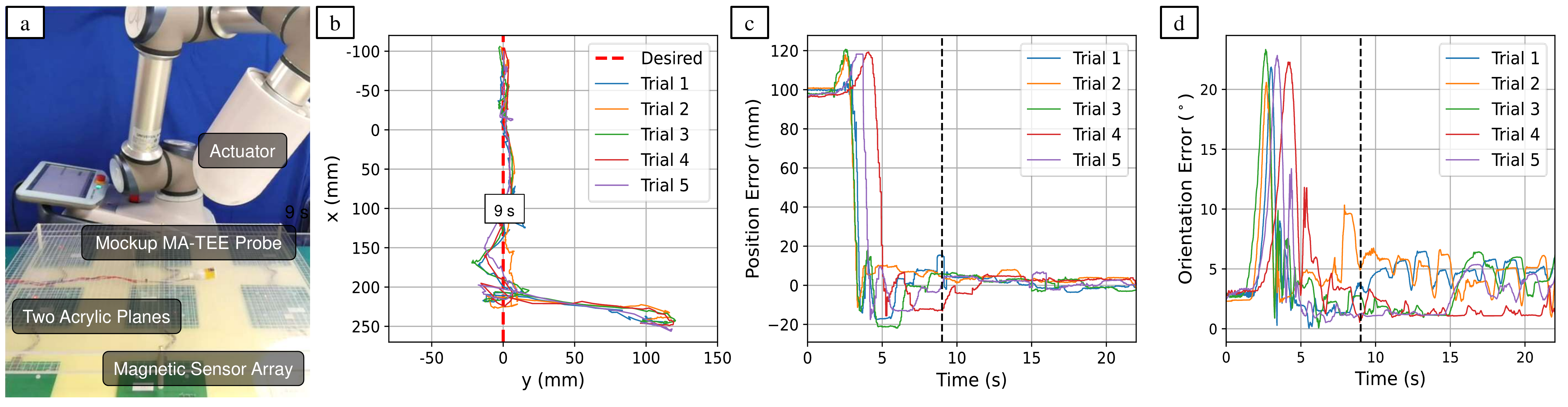}
\caption{(a) shows the experimental setup where the mockup MA-TEE probe was actuated to follow a straight-line trajectory between two acrylic planes placed above the sensor array. (b-d) show the estimated trajectories and tracking errors in 5 trials when the initial position of the probe is offset from the desired trajectory in the $y$ direction by about \SI{100}{\mm}. The red dashed lines in (b) illustrate the desired trajectory, and the black dashed lines in (c)(d) highlight the time when the probe has traveled for \SI{9}{\s}, which is used to calculate the steady-state tracking errors.}
\label{Fig_exp_actuation_quantitative}
\end{figure*}

\begin{figure*}[t]
\centering
\includegraphics[scale=1.0,angle=0,width=0.9\textwidth]{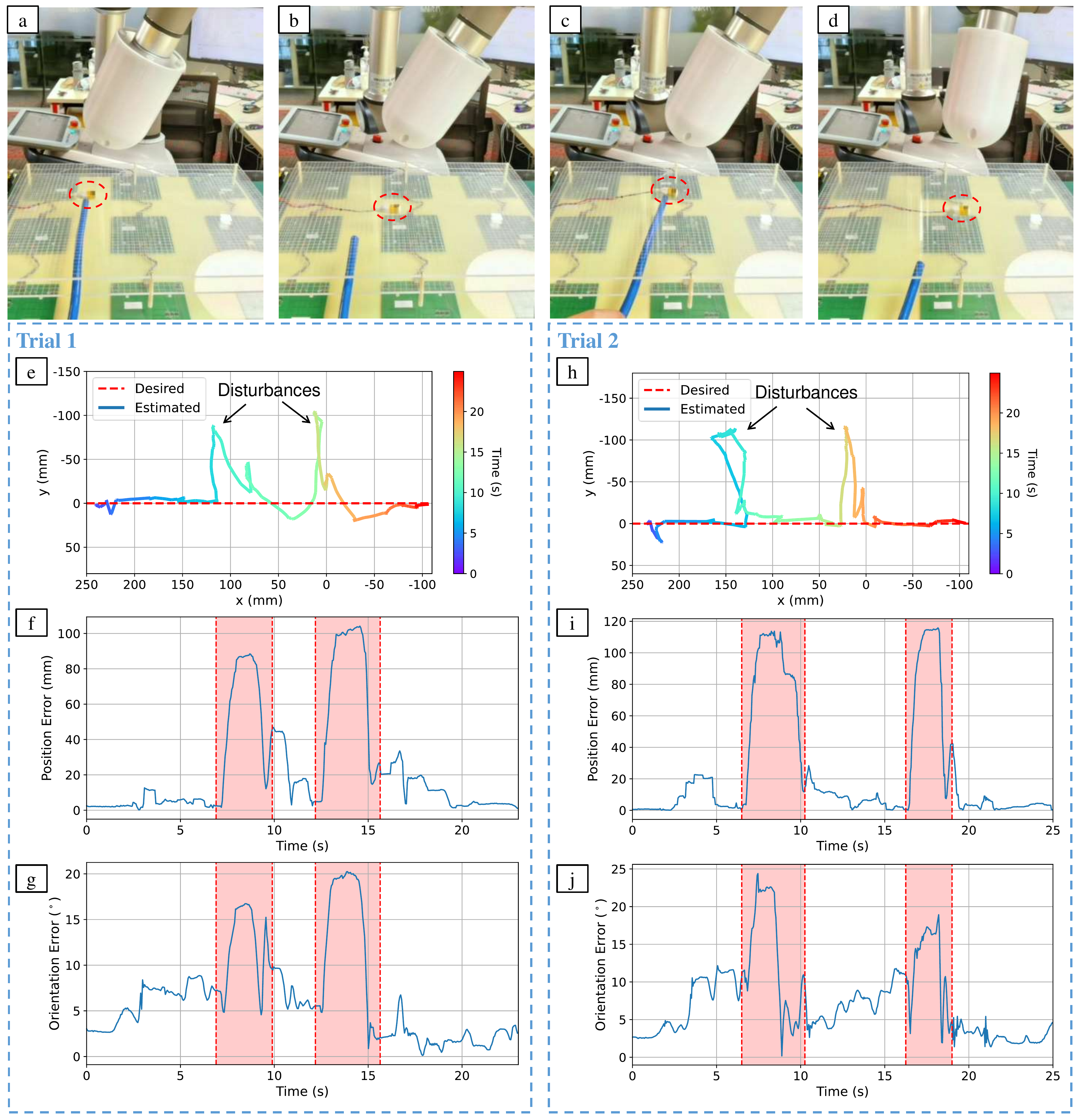}
\caption{(a-d) show the snapshots of the closed-loop control experiments on a straight-line trajectory with two manual disturbances. The red dotted circles highlight the positions of the probe tip. (e-g) and (h-j) show the trajectories of the probe (rainbow) compared with the desired trajectory (red dashed lines), and the position and orientation errors in the two trials, respectively. The red shadowed areas highlight the duration of the disturbances.}
\label{Fig_exp_actuation_disturbance}
\end{figure*}

\subsection{Evaluation of the Closed-Loop Control Method}  \label{section_exp_actuation}

In order to validate the effectiveness of the proposed closed-loop control method, we inserted the mockup TEE probe between two acrylic planes placed horizontally above the sensor array, as shown in Fig. \ref{Fig_exp_actuation_quantitative}(a). The vertical position of the probe was limited by the two planes, which was approximately \SI{100}{\mm} above the sensor array. The system was commanded to actuate the probe to follow a straight-line trajectory on the horizontal plane (i.e., $y=0$). The desired magnetic moment of the probe $\widehat{\mathbf{m}}_{d}$ was set to be aligned with the $-z$ direction in the world frame. The acrylic planes were rigidly attached to the sensor array during the experiments. 
The tracking accuracy was represented by the error between the desired trajectory and the localization results given by the pose estimation algorithm. 

First, we conducted 5 trials to assess the control performance of the system when the initial position of the probe deviates from the desired trajectory in the $y$ direction by approximately \SI{100}{mm}. The tracking error in position was computed as the lateral offset from the  trajectory, and the orientation error was computed as the angle between the desired and estimated magnetic moments of the probe $\widehat{\mathbf{m}}_{d}$ and $\widehat{\mathbf{m}}_{c}$. 
Fig. \ref{Fig_exp_actuation_quantitative}(b-d) show the estimated trajectories of the probe compared with the desired trajectory and the tracking errors in position and orientation in the 5 trials. The steady-state position error after the probe has traveled for \SI{9}{\s} was $1.97 \pm 2.97$ \si{mm}, and the steady-state orientation error was $3.63 \pm 1.75 ^\circ$ over the 5 trials. 

We additionally conducted two experiments to demonstrate the capability of the system to overcome undesired disturbances and maintain a small steady-state tracking error. As shown in Fig. \ref{Fig_exp_actuation_disturbance}, the probe was initialized on the desired trajectory, and two manual disturbances were implemented during the voyage of the probe to push the probe tip laterally by approximately \SI{100}{\mm} when it has travelled about \SI{120}{\mm} and \SI{240}{\mm}. It can be seen from Fig. \ref{Fig_exp_actuation_disturbance} (b)(d) that the system quickly responded to the position errors and controlled the probe to recover from the disturbances. Fig. \ref{Fig_exp_actuation_disturbance} (e-g) and (h-j) show the estimated moving trajectories and tracking errors in the two trials, respectively. The steady tracking errors calculated over the trajectory before and after the two disturbances were $3.23 \pm \SI{4.80}{mm}$  and $3.86 \pm \ang{2.40}$ for the first experiment, and $2.59 \pm \SI{7.93}{mm}$ and 
$4.80  \pm \ang{3.36}$ for the second experiment, respectively.

\begin{figure*}[tb]
\centering
\includegraphics[scale=1.0,angle=0,width=0.99\textwidth]{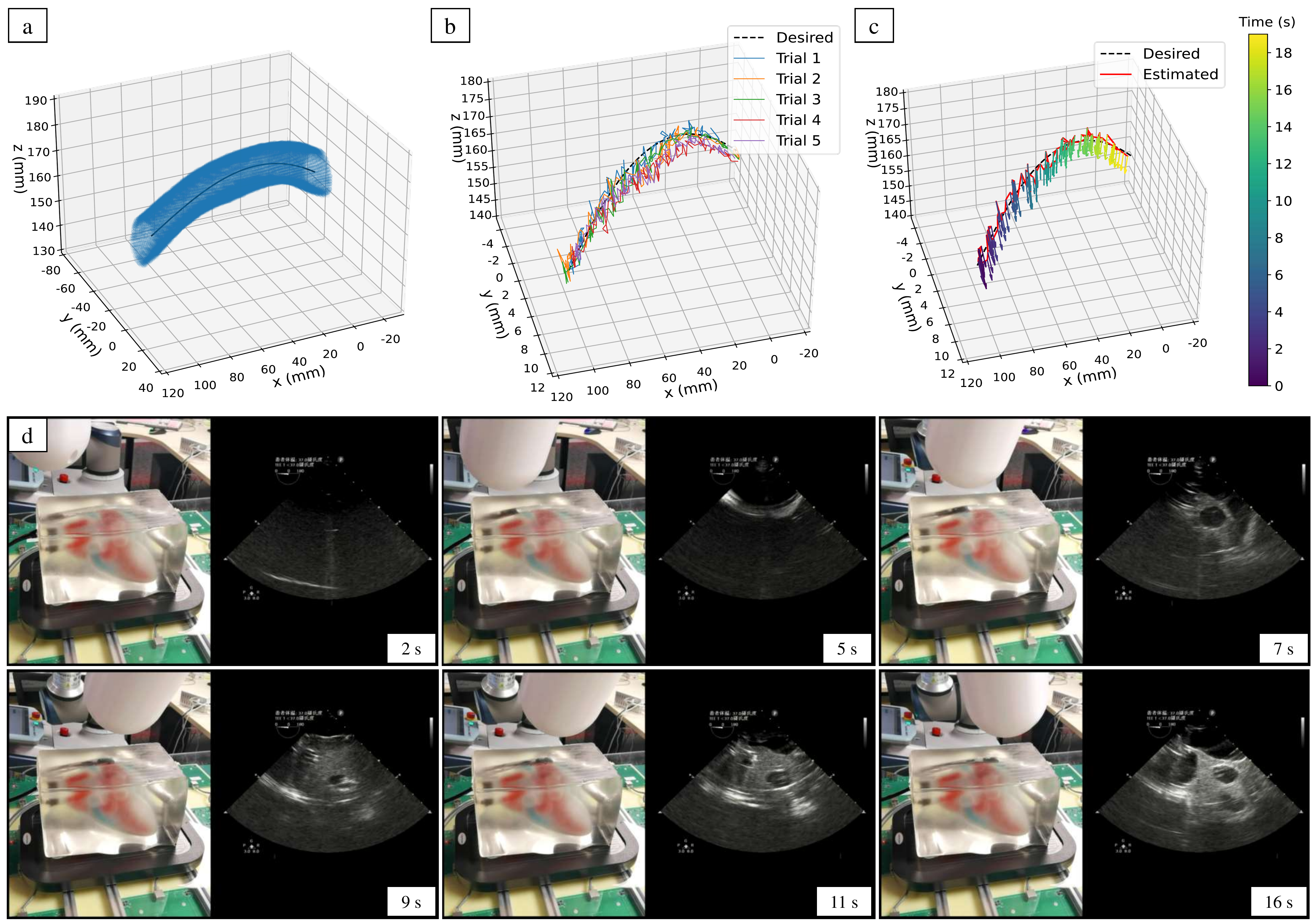}
\caption{(a) shows the esophageal points (blue) and the fitted centerline trajectory of the esophagus (black). (b) shows the estimated 3D trajectories of the probe in all the propulsion experiments compared with the desired trajectory (black dashed line). (c) Arrows illustrate the $z$-direction of the probe in one trial, with the color gradation showing the progression of time. (d) shows some snapshots of the experimental setup and the corresponding ultrasound images during the propulsion experiments in the cardiac phantom.}
\label{Fig_exp_phantom_advance}
\end{figure*}

\begin{figure*}[tb]
\centering
\includegraphics[scale=1.0,angle=0,width=0.7\textwidth]{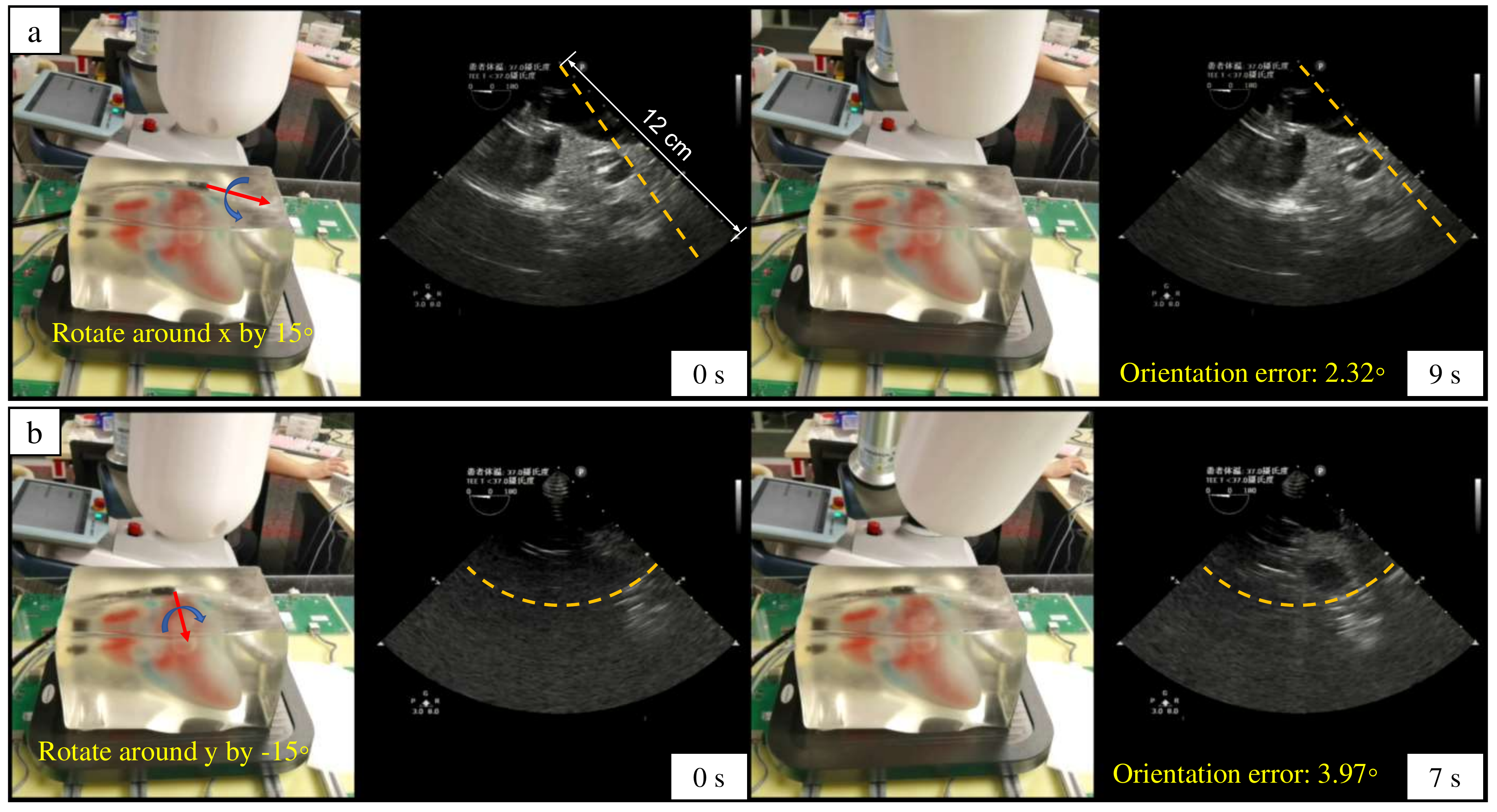}
\caption{(a) and (b) illustrate the in-plane rotation and out-of-plane rotation of the MA-TEE probe in the cardiac phantom. The red and blue arrows show the rotation axes and directions in the two experiments, respectively. The yellow dotted lines in the ultrasound images are used to highlight the differences between the images acquired before and after the rotations. The imaging depth is marked in the first ultrasound image (white). The commanded rotation angles and the final orientation errors in the two experiments are marked in yellow.}
\label{Fig_exp_phantom_rotation}
\end{figure*}

\subsection{Demonstration of Robotic MA-TEE in a Cardiac Phantom}
To preliminarily demonstrate the potential of the system in robotic TEE acquisitions, we further validated our proposed MA-TEE system on a realistic echocardiography training phantom with simulated cardiac tissues (Blue Phantom Cardiac Echo Model BPH700, CAE Healthcare), as shown in Fig. \ref{Fig_system_setup}(a). The TEE transducer has a field of view of \ang{90} and the imaging depth was set to \SI{120}{\mm}. The simulated esophagus was lubricated with ultrasound coupling gel to ensure acoustic coupling and improve the lubricity of the probe-tissue interface. 

In the first set of experiments, the MA-TEE probe was commanded to advance through the esophagus following a given trajectory in the cardiac phantom in 5 trials. To obtain the trajectory of the esophagus, we performed a CT scan of the phantom, and asked a clinician to manually segment the esophagus from the CT image. Then, polynomial fitting and B-spline interpolation were performed on the esophageal points to obtain a smooth esophageal trajectory, which was then transferred to the world coordinate system as the desired trajectory for magnetic control, as shown in Fig. \ref{Fig_exp_phantom_advance}(a). The desired $z$-direction of the probe was set to be perpendicular to the tangential direction of the esophagus and located in the vertical plane.
In all trials, the electronic steering angle was set as 0, which means the imaging plane coincided with the $y$-$z$ plane of the probe.

The estimated  3D trajectories in all five trials are shown in Fig. \ref{Fig_exp_phantom_advance}(b), and Fig. \ref{Fig_exp_phantom_advance}(c) illustrates the $z$-direction of the probe along the trajectory in one trial. Snapshots of the experimental setup and the corresponding ultrasound images are shown in Fig. \ref{Fig_exp_phantom_advance}(d). The total length of the moving trajectory of the probe was about \SI{200}{\mm}, and the average propulsion time in all trials was \SI{20}{\s}. The overall tracking accuracy in all five trials was $5.40 \pm \SI{3.21}{mm}$  and $7.84 \pm \ang{4.73}$ in position and orientation, respectively. The results demonstrated the effectiveness of the system to smoothly control the probe to advance in the narrow esophagus with variable friction and acquire clear ultrasound images of the heart.
The separation distance between the actuator and the probe was about \SI{200}{\mm}, generating a maximum magnetic coupling force of \SI{0.85}{\N} applied to the probe tip, which was empirically found sufficient for actuation.

In the second test, we qualitatively demonstrate the in-plane and out-of-plane rotations of the probe performed by the MA-TEE system in the cardiac phantom under tele-operation, as shown in Fig. \ref{Fig_exp_phantom_rotation}. First, the probe was required to rotate around its $x$-axis by \ang{15}. The images acquired before and after applying the rotation are shown in Fig. \ref{Fig_exp_phantom_rotation}(a). 
Then, the probe was commanded to perform an out-of-plane rotation around its $y$-axis by \ang{-15}, which corresponds to the ``anteflex” operation in traditional TEE manipulation. As illustrated in Fig. \ref{Fig_exp_phantom_rotation}(b), the heart cavity was not visualized with the original probe orientation, and was clearly visualized after the out-of-plane rotation executed by the MA-TEE system. 
These results have demonstrated the capability of the proposed MA-TEE system to adjust the position and orientation of the probe under remote control for TEE image acquisitions.

\subsection{Video Demonstration}
A detailed video demonstration of our experiments in Section~V-B to V-D is provided in the supplementary multimedia attached to this paper, and also available online at the following links. An illustration of the localization experiments is given in Supplementary Video 1\footnote{\url{https://youtu.be/dynumUn6WAU}}. An illustration of the closed-loop control experiments is provided in Supplementary Video 2\footnote{\url{https://youtu.be/sNbxjVdleNU}}. Our demonstration of robotic TEE in the cardiac phantom can be found in Supplementary Video 3\footnote{\url{https://youtu.be/wMU_8lsNEeI}}.

\section{Conclusions}
In this paper, we have presented the first closed-loop magnetic manipulation framework to perform 6-DOF pose estimation and 5-DOF control of a robotic TEE probe. We have shown that by modifying a standard TEE probe to forgo the flexible gastroscope and attach the probe tip with a magnet and an IMU sensor, the probe can be viewed as a magnetic capsule robot, and direct manipulation of the distal tip of the probe in the esophagus can be achieved based on an external and internal sensor fusion approach. Our method can achieve accurate and efficient control of the magnetic capsule robot without requiring complex structures of the capsule or the external actuator, and can circumvent the need for applying specific motions of the capsule.
Our results show that 
\begin{enumerate}
\item Using an external magnetic sensor array and an internal IMU, the proposed EKF-based localization method can achieve fast and accurate pose estimation of the probe tip in a large workspace of  \SI[parse-numbers=false]{0.5 $\times$ 0.5 $\times$ 0.2}{\cubic\m} at an update rate of $80 \sim \SI{90}{\hertz}$, and is compatible with simultaneous magnetic actuation.
\item The proposed magnetic localization and actuation methods can realize closed-loop control of the probe tip to follow a manually specified trajectory in a 3D workspace.
\item  We provide the first demonstration of a magnetically manipulated, tele-operated TEE probe for ultrasound acquisitions. The results on a cardiac imaging phantom demonstrate the potential of the proposed framework to be applied in real conditions.
\end{enumerate}

The MA-TEE probe prototype used in this work was manually assembled and fabricated to preliminarily validate the feasibility of the proposed approach. In view of a clinical translation, the design of the probe will need further refinements, e.g., to use a more spherical shape to reduce friction, and coat the probe with soft and biocompatible materials \cite{norton2019intelligent}. 
Moreover, the MA-TEE probe and the soft tether can be further miniaturized by integrating the TEE transducer and the magnetic control modules in a more compact design. The size of the permanent magnet in the TEE probe can be reduced by increasing the volume of the actuator magnet to keep the same actuation distance \cite{mahoney2016five}. It should also be noted that our experimental results only provide a preliminary validation of the closed-loop control system for tele-operated TEE acquisitions without considering the dynamic disturbances that may occur in a clinical setting, such as respiration and heartbeat. In vivo trials (e.g., in a porcine model) and clinical tests will be necessary to further demonstrate the effectiveness of the methods toward a potential clinical translation.

Maintaining close contact between the probe and the esophageal tissue is crucial for acoustic coupling and effective imaging for TEE acquisitions. Adding more coupling gel on the probe could temporarily solve this issue, but cannot guarantee close probe-tissue contact during the movement. In view of this, visual servoing methods that control the pose of the probe based on the ultrasound image quality may improve contact between the probe and the tissue for better image quality \cite{chatelain2017confidence}. 
The magnetic control method itself may also be exploited to improve transducer coupling by placing the actuator in front of the patient so that the magnetic attraction will inherently provide close contact between the probe and the tissue for effective imaging of the heart \cite{norton2019intelligent}.

It is challenging to move the transducer backward using the current control method due to the large resistance caused by the tether. Therefore, our method is mainly proposed to propel the transducer forward in the esophagus and steer it for the examination, and the tether may be manually retrieved by a human operator during or after the examination, which can also be used as a failsafe method, similar to \cite{norton2019intelligent}. In view of a teleoperation scenario where trained clinicians might not be present, one may consider using another mechanism to grab and hold the cable to pull the transducer out \cite{wang2016design}. Another possible solution is to replace the tether with wireless technology \cite{seo2017ultrasound}. 
In addition, it might be useful in clinical practice to still have a mechanism to allow manual insertion of the probe tip to handle the situation where magnetic propulsion may not be sufficient to overcome environmental resistance. 

Nevertheless, the proposed methods have provided a novel solution to closed-loop control of an intracorporeal ultrasound probe based on magnetic methods. In view of a fully autonomous robotic system for TEE acquisition, the control method presented in this paper may be integrated with image-guided probe navigation methods (e.g., \cite{li2023rltee, li2021image}). The proposed methods may also be extended to other applications such as robot-assisted minimally invasive surgeries \cite{10058160}.

\bibliographystyle{IEEEtran}  
\bibliography{bare_jrnl}
\section*{Acknowledgments}
We would like to thank Dr. Chao Hu, Mr. Quan Yue, Ms. Yue Meng, and Dr. Songhua Xiao from Yuanhua Robotics, Perception \& AI Technologies Ltd., Shenzhen, China, for their inspiring discussion and technical support to this study.
%
%
%


\newpage


\begin{IEEEbiography}[{\includegraphics[width=1in,height=1.25in,clip,keepaspectratio]{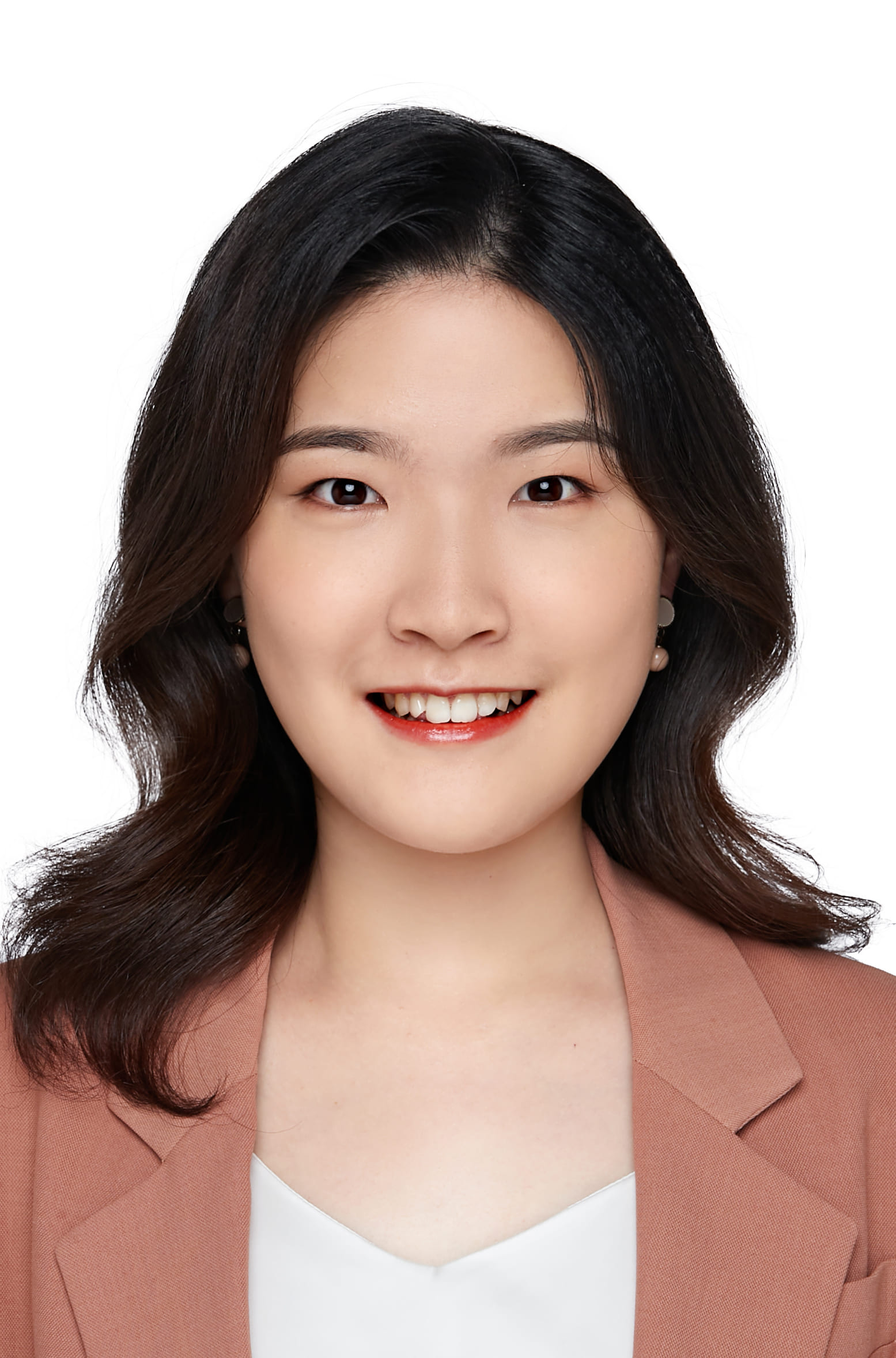}}]
{Keyu Li} received the B.Eng. degree in communication engineering from Harbin Institute of Technology, Weihai, China, in 2019. She is currently pursuing the Ph.D. degree with the Department of Electronic Engineering, The Chinese University of Hong Kong, Hong Kong. 

Her research interests include artificial intelligence in robot decision-making, medical robotics, and medical imaging applications, with a focus on autonomous robotic ultrasound systems, supervised by Prof. Max Q.-H, Meng.

Ms. Li received the IEEE Transactions on Automation Science and Engineering Best New Application Paper Award, in 2022. She has been an awardee of the Hong Kong PhD Fellowship Scheme (HKPFS) since 2019. 
\end{IEEEbiography}

\begin{IEEEbiography}[{\includegraphics[width=1in,height=1.25in,clip,keepaspectratio]{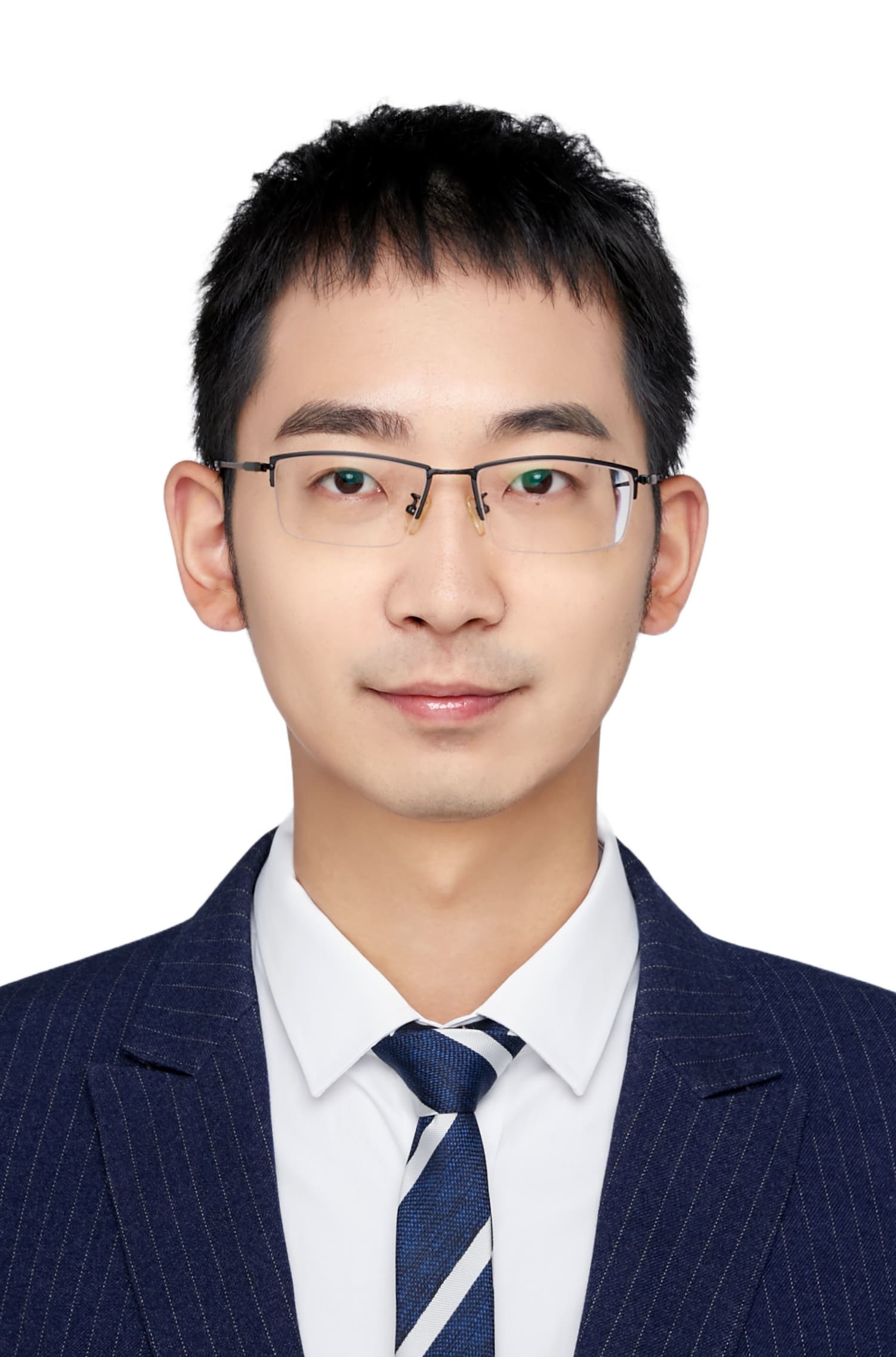}}]{Yangxin Xu}

received the B.Eng. degree in electrical engineering and its automation from Harbin Institute of Technology, Weihai, China, in 2017, and the Ph.D. degree from the Department of Electronic Engineering, The Chinese University of Hong Kong, Hong Kong SAR, China, in 2021. He is currently with Yuanhua Robotics, Perception \& AI Technologies Ltd., Shenzhen, China.

His research focuses on magnetic actuation and localization methods and hardware implementation for active wireless robotic capsule endoscopy, supervised by Prof. Max Q.-H, Meng.

Dr. Xu received the IEEE Transactions on Automation Science and Engineering Best New Application Paper Award, in 2022, and the Best Conference Paper Award from the 2018 IEEE International Conference on Robotics and Biomimetics (ROBIO), Kuala Lumpur, Malaysia, in 2018.

\end{IEEEbiography}

\begin{IEEEbiography}[{\includegraphics[width=1in,height=1.25in,clip,keepaspectratio]{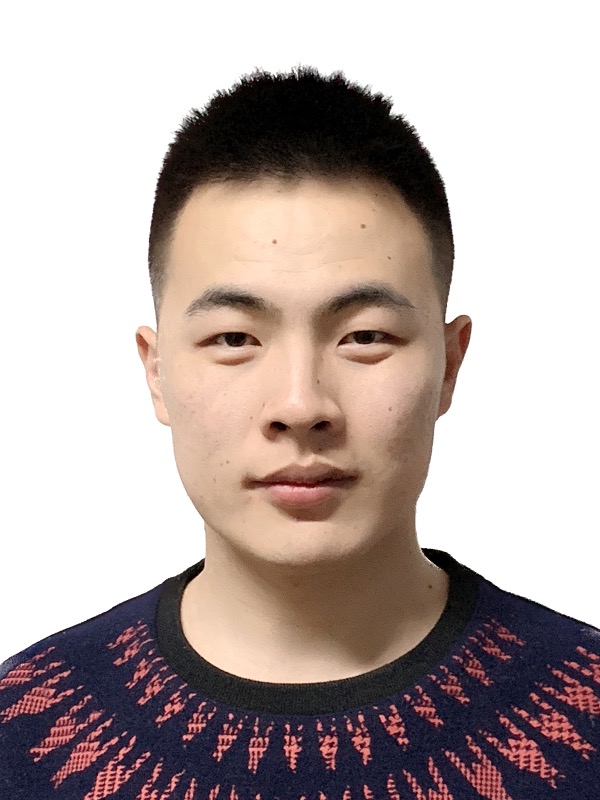}}]
{Ziqi Zhao} received the B.Eng. and M.Eng. degree in mechanical engineering from Shenyang Jianzhu University, Shenyang, China, in 2016 and 2019, respectively. He is currently pursuing the Ph.D. degree with the Department of Electronic and Electrical Engineering, the Southern University of Science and Technology (SUSTech), Shenzhen, China.

His current research interests include bionic, medical and service robotics, supervised by Prof. Max Q.-H, Meng.

Mr. Zhao received the IEEE Transactions on Automation Science and Engineering Best New Application Paper Award, in 2022.

\end{IEEEbiography}

\begin{IEEEbiography}[{\includegraphics[width=1in,height=1.25in,clip,keepaspectratio]{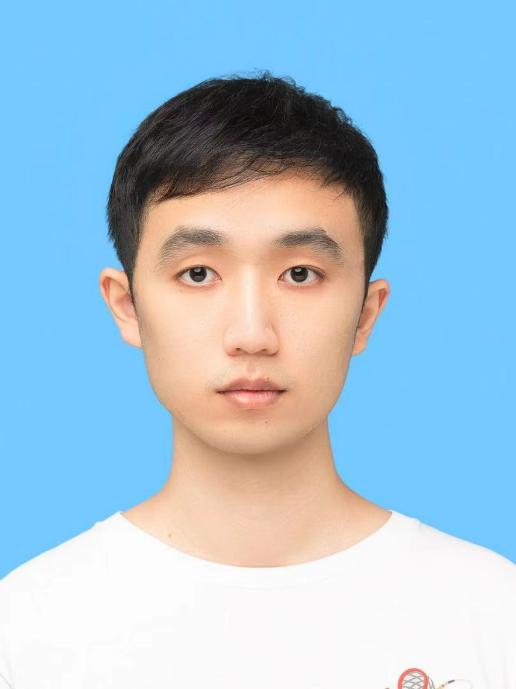}}]
{Ang Li} received his B.S. in biomedical engineering from Northeastern University, China, in 2019 and MSc. degree in electronic engineering from the Chinese University of Hong Kong, in 2020. Currently, he is a Ph.D. student in the electronic engineering department at CUHK and his research interests include image-based surgical navigation and reinforcement learning. 

\end{IEEEbiography}

\begin{IEEEbiography}[{\includegraphics[width=1in,height=1.25in,clip,keepaspectratio]{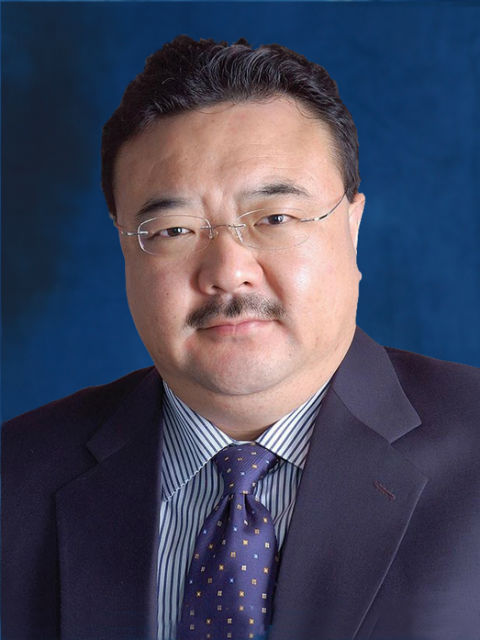}}]
{Max Q.-H. Meng} received his Ph.D. degree in Electrical and Computer Engineering from the University of Victoria, Canada, in 1992. He is currently a Chair Professor and the Head of the Department of Electronic and Electrical Engineering at the Southern University of Science and Technology in Shenzhen, China, on leave from the Department of Electronic Engineering at the Chinese University of Hong Kong. He joined the Chinese University of Hong Kong in 2001 as a Professor and later the Chairman of Department of Electronic Engineering. He was with the Department of Electrical and Computer Engineering at the University of Alberta in Canada, where he served as the Director of the ART (Advanced Robotics and Teleoperation) Lab and held the positions of Assistant Professor (1994), Associate Professor (1998), and Professor (2000), respectively. He is an Honorary Chair Professor at Harbin Institute of Technology and Zhejiang University, and also the Honorary Dean of the School of Control Science and Engineering at Shandong University, in China. 

His research interests include medical and service robotics, robotics perception and intelligence. He has published more than 750 journal and conference papers and book chapters and led more than 60 funded research projects to completion as Principal Investigator. 

Prof. Meng has been serving as the Editor-in-Chief and editorial board of a number of international journals, including the Editor-in-Chief of the Elsevier Journal of Biomimetic Intelligence and Robotics, and as the General Chair or Program Chair of many international conferences, including the General Chair of IROS 2005 and ICRA 2021, respectively. He served as an Associate VP for Conferences of the IEEE Robotics and Automation Society (2004-2007), Co-Chair of the Fellow Evaluation Committee and an elected member of the AdCom of IEEE RAS for two terms. He is a recipient of the IEEE Millennium Medal, a Fellow of IEEE, a Fellow of Hong Kong Institution of Engineers, and an Academician of the Canadian Academy of Engineering.

\end{IEEEbiography}


\end{document}